\documentclass[11pt]{article}

\usepackage[preprint]{acl}

\usepackage{times}
\usepackage{latexsym}

\usepackage[T1]{fontenc}

\usepackage[utf8]{inputenc}

\usepackage{microtype}

\usepackage{inconsolata}

\usepackage{graphicx}

\usepackage{caption}
\usepackage{subcaption}
\usepackage{array}
\usepackage{longtable}
\usepackage{pdflscape}
\usepackage{booktabs}
\usepackage{multirow}
\usepackage{algorithm}
\usepackage{algorithmic}
\usepackage{amsmath}
\usepackage{amssymb}
\usepackage{amsthm}
\usepackage{makecell}

\newtheoremstyle{obsstyle}%
  {1ex}{1ex}{\itshape}{}{\bfseries}{.}{1em}{}
\theoremstyle{obsstyle}

\title{N-GLARE: An Non-Generative Latent Representation-Efficient LLM Safety Evaluator}

\author{
  \textbf{Zheyu Lin\textsuperscript{1}},
  \textbf{Jirui Yang\textsuperscript{2}},
  \textbf{Yukui Qiu\textsuperscript{3}}
\\
  \textbf{Hengqi Guo\textsuperscript{2}},
  \textbf{Yubing Bao\textsuperscript{2}},
  \textbf{Yao Guan\textsuperscript{2}}
\\
\\
  \textsuperscript{1}University of California, Riverside,
  \textsuperscript{2}Fudan University,
  \textsuperscript{3}South China University of Technology
\\
  \small{
Zheyu Lin: \href{mailto:1298lzy@gmail.com}{1298lzy@gmail.com}\quad
Jirui Yang: \href{mailto:yangjr23@m.fudan.edu.cn}{yangjr23@m.fudan.edu.cn}\quad
Yukui Qiu: \href{mailto:202162311356@mail.scut.edu.cn}{202162311356@mail.scut.edu.cn}}
\\
\small{
Hengqi Guo: \href{mailto:hqguo22@m.fudan.edu.cn}{hqguo22@m.fudan.edu.cn}\quad
Yubing Bao: \href{mailto:ybbao23@m.fudan.edu.cn}{ybbao23@m.fudan.edu.cn}\quad
Yao Guan: \href{mailto:yguan25@m.fudan.edu.cn}{yguan25@m.fudan.edu.cn}
}
}

\begin{document}
\maketitle
\begin{abstract}
Evaluating the safety robustness of LLMs is critical for their deployment. However, mainstream Red Teaming methods rely on online generation and black-box output analysis. These approaches are not only costly but also suffer from feedback latency, making them unsuitable for agile diagnostics after training a new model.
To address this, we propose N-GLARE (A Non-Generative, Latent Representation-Efficient LLM Safety Evaluator). N-GLARE operates entirely on the model's latent representations, bypassing the need for full text generation. It characterizes hidden layer dynamics by analyzing the APT (Angular-Probabilistic Trajectory) of latent representations and introducing the JSS (Jensen-Shannon Separability) metric.
Experiments on over 40 models and 20 red teaming strategies demonstrate that the JSS metric exhibits high consistency with Red Teaming safety rankings at less than 1\% token and runtime cost.

\end{abstract}

\section{Introduction}

Large Language Models must be aligned to ensure their behavior consistently follows human values and safety norms, rather than merely maximizing task performance \cite{das2025security,dong2024attacks}. Without such alignment, models may generate harmful, biased, or unsafe outputs when facing adversarial prompts or complex contexts \cite{wei2023jailbroken,russinovich2025great,yang2025concept}.

Evaluating model safety is a foundational step toward achieving such alignment. The dominant paradigm in both industry and academia is currently Red Teaming \cite{lin2025against}, which simulates adversarial user behavior by designing diverse, subtle attack prompts to probe the model and elicit unsafe outputs. While this online, behavior-based testing has proved effective for discovering vulnerabilities, it suffers from two fundamental limitations. First, it is highly resource-intensive: test coverage depends on the scale and diversity of attack prompts, which require continual human and computational investment as new attack vectors emerge. Second, Red Teaming is essentially a black-box methodology. It only observes input--output behavior and provides little insight into the model’s \emph{internal} safety state. A refusal that looks safe at the surface level may simply be triggered by brittle pattern matching, rather than reflecting a robust internal representation of harmfulness.

To move beyond purely behavioral stress testing, recent work analyzes LLMs in their latent representation space \cite{zou2023representation,li2025revisiting}. Representation engineering shows that different semantic concepts—such as truthfulness, refusal, or harmfulness—correspond to structured patterns in hidden activations, and contrasting concept sets can be distinguished by consistent directions in this space, suggesting that high-level semantics are explicitly encoded in latent geometry. 

Existing representation analysis work primarily uses such directions at the \emph{prompt level}, for example to steer individual generations or to classify the risk of a specific input. However, these techniques are typically model-specific and do not directly provide a \emph{model-level} safety indicator that is comparable across architectures and checkpoints.

In this paper, we take a step in that direction and ask: \textbf{Can we evaluate model safety without generation, using internal representations that correlate with red-teaming robustness?
}

To this end, we propose \textbf{N-GLARE} (\emph{A Non-Generative, Latent Representation-Efficient LLM Safety Evaluator}). For each model, we construct \textbf{Angular-Probabilistic Trajectories} (APT) of hidden representations under different probing conditions: benign ($B$), jailbreak ($J$), plainquery ($P$) and an idealized refusal counterfactual ($R$). After normalizing these trajectories along a standardized progress axis, we model benign activations as a low-dimensional manifold and define a geometric \emph{turning angle} that measures how strongly a trajectory deviates away from this benign manifold over time. We then introduce \textbf{Jensen--Shannon Separability} (JSS), which compares the slice-wise distributions of these turning angles across conditions, and aggregates them into model-level safety indicators, introducing other JSS-based safety scores.

\begin{figure}
    \centering
    \scalebox{1.0}{%
    \includegraphics[width=\linewidth]{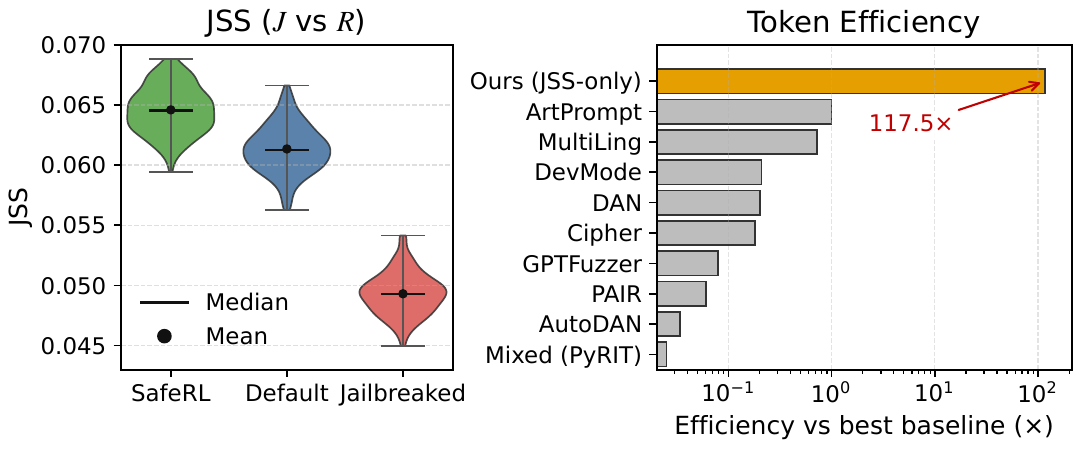}
    }
    \caption{
    JSS exhibits clear separability across safety states while achieving orders-of-magnitude higher token efficiency than red-teaming baselines.
    }
    \label{fig:1}
\end{figure}

\textbf{The core intuition is that: internal trajectory separability can serve as a proxy for the underlying safety state of the model, potentially reproducing the discriminative power of large-scale Red Teaming at a fraction of the cost.}

Figure~\ref{fig:1} supports the core intuition: when comparing multiple variants of the same base model (e.g., RL-aligned, base, and safety-removed versions of Qwen3-4B), we observe that better-aligned variants exhibit more pronounced geometric separation between different trajectories in latent space. 

Experiments on over 40 models and 20 red-teaming strategies demonstrate that N-GLARE faithfully reproduces the safety rankings induced by traditional Red Teaming while using less than 1\% of the token and runtime cost, making it suitable for agile diagnostics after training new models or updating safety mechanisms.

Our main contributions are summarized as follows:
\begin{itemize}
    \item \textbf{A non-generative, latent-space safety evaluator.}
    We propose N-GLARE, a framework that entirely bypasses online text generation and external output evaluators. To our knowledge, this is the first work to leverage latent representation trajectories for \emph{model-level} LLM safety assessment.

    \item \textbf{A geometric formalization of safety alignment.}
    We formulate safety evaluation as a trajectory separability problem in representation space, construct a benign manifold via whitened PCA, and define a introduce Angular-Probabilistic Trajectories (APT) deviation measure. Building on this, we propose Jensen--Shannon Separability (JSS) and JSS-related normalized safety score as statistically robust, cross-model comparable indicators.

    \item \textbf{Extensive empirical validation and efficiency gains.}
    Across more than 40 models and 20 red-teaming strategies, we show that JSS-based rankings are highly consistent with those derived from traditional Red Teaming, while reducing token and runtime cost to below 1\%. We further demonstrate that N-GLARE supports fine-grained analyses such as early-stage safety dynamics and comparison of alignment interventions across checkpoints, providing a practical tool for real-time safety diagnostics.
\end{itemize}

\section{Technical Background}
\subsection{LLM Security and Red Teaming}

Red teaming evaluates LLM safety by simulating attacker strategies to proactively identify and probe model vulnerabilities.Existing Red Teaming approaches can be broadly categorized into two types. The first focuses on content safety in single-turn interactions, such as leveraging algorithms like Monte Carlo Tree Search (MCTS) and evolutionary algorithms to optimize attack prompts \cite{sadasivan2024fast}, or generating jailbreak templates through the mutation of seed prompts \cite{yu2023gptfuzzer}. The second category addresses security risks in multi-turn dialogue scenarios, typically employing specialized agents for attack planning, executing dynamic multi-turn jailbreaks, and evaluating the attack's efficacy \cite{rahman2025x,weng2025foot,ren2025llms}.

Existing methods, despite increasing diversification, largely follow a three-stage pipeline that includes attack prompt construction, target model generation, and output-based evaluation, which inherently incurs high overhead in red-teaming assessments.

\subsection{LLM Latent Space Analysis Techniques}

The internal mechanisms of LLMs are often treated as a black box. To understand how they process information, researchers analyze their latent space, i.e., the internal high-dimensional hidden states. In Transformers, this corresponds to the activation vector $h_{l, p} \in \mathbb{R}^d$ at a specific layer $l$ and token position $p$, where $d$ is the hidden dimension.

\subsubsection{Embedding Analysis}
Embedding analysis reveals semantic meaning by examining the static geometry of latent space vectors. Common methods like cosine similarity ($Sim(\vec{A}, \vec{B}) = \frac{\vec{A} \cdot \vec{B}}{\|\vec{A}\| \|\vec{B}\|}$) and dimensionality reduction visualization like t-SNE can reveal semantic clusters (e.g., ``king" with ``queen") \cite{rogers2020primer}. Furthermore, domain-adaptation analysis shows that fine-tuning such as BioBERT \cite{lee2020biobert} reshapes the latent space, forming more refined domain-specific subspaces to enhance downstream task performance.

\subsubsection{Probing and Intervention}

This technique shifts from passive analysis to active decoding and control. Early probing \cite{tenney2019bert} used simple linear classifiers ($y = f(H)$) to test if latent activations $H$ encoded specific linguistic properties, discovering that information is stored hierarchically in LLMs.

A more advanced paradigm, Representation Engineering (RepE) \cite{zou2023representation}, controls model behavior by directly manipulating latent representations. Its core is identifying vector directions, or concept operators $o^c$, that represent high-level concepts (e.g., truthfulness). This is often achieved by contrasting the activation differences from different input sets ($\mathcal{S}^{c+}$ vs $\mathcal{S}^{c-}$): $o^c \approx \mathbb{E}[H(\mathcal{S}^{c+})] - \mathbb{E}[H(\mathcal{S}^{c-})]$
Here, $H(\mathcal{S})$ represents the mean activation of model $\mathcal{M}$ when processing a specific input set $\mathcal{S}$. RepE is widely applied in LLM safety \cite{cai2024self, wang2025model, wang2024inferaligner, cao2024nothing}, for instance, by using positive or negative steering vectors to enhance truthfulness \cite{marks2023geometry, li2023inference, wang2025adaptive, rimsky2024steering} or suppress meta-behaviors like ``refusal" \cite{arditi2024refusal}.

\subsubsection{Latent Space Dynamic Analysis}
Dynamic analysis focuses on the activation trajectories during the reasoning process. Researchers not only track attention changes during Chain-of-Thought (CoT) \cite{wei2022chain} but also begin to analyze the latent space trajectories themselves. For example, Li et al. \cite{li2025core} proposed Chain-of-Reasoning Embedding (CoRE). They found that redundant reasoning manifests as identifiable cyclical fluctuations in the trajectory. Based on this, it enables label-free self-evaluation to dynamically terminate inefficient reasoning, thereby improving efficiency.

Despite significant progress in analysis techniques for LLM latent spaces, the field still faces the challenge of model-specificity: the identified features, concept operators (like $o^c$), attention heads, or reasoning trajectories are often highly dependent on a single model $\mathcal{M}$ and lack transferability across models.

\section{Methodology}
\label{sec:method}
This section formalizes the representation-space safety evaluation framework proposed above. Our goal is to map internal activation trajectories to scalar safety metrics. We first define the notation and probing conditions, followed by the framework's three core stages: (i) trajectory extraction and standardization; (ii) benign manifold fitting and geometric turning angle calculation; and (iii) distribution divergence measurement and safety score aggregation.

\subsection{Preliminaries: Symbols and Probing Conditions} \label{sec:preliminaries}
Given a model set $\mathcal{M} = \{M_1,\dots,M_K\}$, we analyze internal representation behaviors under varying inputs. We define the probing condition $A \in \{B, J, R, P\}$ to correspond to four interaction history types:

\textbf{Benign ($B$):} Standard task instructions conforming to natural distributions, used to fit the benign reference manifold.

\textbf{Jailbreak ($J$):} Adversarial sequences designed to induce harmful outputs.

\textbf{Ideal Refusal ($R$):} Identical to $J$ in user content (retaining harmful intent) but forcing the assistant response to a policy-compliant refusal template. This represents the model's ideal defensive state against attacks.

\textbf{Plain Query ($P$):} Direct presentation of harmful intent (without disguise) to capture the model's unconstrained internal response to harmful information.

To enhance cross-model comparability and reduce dimensionality, we partition Transformer layers into groups $G \in \{\text{lower}, \text{middle}, \text{upper}\}$ and average within each group to obtain group-level representations $h^{(A,G)}$.

\subsection{Stage 1: Trajectory Extraction and Standardization}
\label{sec:trajectory_and_standard}

For a model $M$ and condition $A$, we construct probing sequences. At each time step $t$, we extract the hidden states of the last token across all layers at the decision-making cross-section immediately preceding the next assistant response generation:

\begin{equation}
\Phi_M^{(A)}(t) = \bigl(h_{L,t}^{(A)}, \dots, h_{1,t}^{(A)}\bigr), \quad h_{\ell,t}^{(A)} \in \mathbb{R}^{d_\ell},
\end{equation}

where $L$ is the number of layers and $d_\ell$ is the hidden dimension. As defined, these are aggregated into group-level representations $h_t^{(A,G)} \in \mathbb{R}^{d}$. Thus, for condition $A$ and group $G$, the sequence $\{h_t^{(A,G)}\}_{t=1}^{T_A}$ defines a trajectory in latent space.

Since generation lengths $T_A$ vary across prompts and models, we define an arc-length standardized progress variable $s_t \in [0,1]$ to align trajectories on a unified timeline. For $t > 1$ (with $s_1 = 0$):

\begin{equation}
s_t = \frac{\sum_{k=1}^{t-1} \Vert h_{k+1}^{(A,G)} - h_k^{(A,G)} \Vert_2}{\sum_{k=1}^{T_A-1} \Vert h_{k+1}^{(A,G)} - h_k^{(A,G)} \Vert_2}
\label{eq:progress}
\end{equation}

This parameterization treats trajectories as curves evolving along a shared progress axis $s$. The standardized trajectory is denoted as:

\begin{equation}
\Gamma_M^{(A,G)} = {(h_t^{(A,G)}, s_t)}_{t=1}^{T_A}.
\end{equation}








\begin{figure}[t]
    \centering
    \scalebox{0.75}{
    \includegraphics[width=1\linewidth]{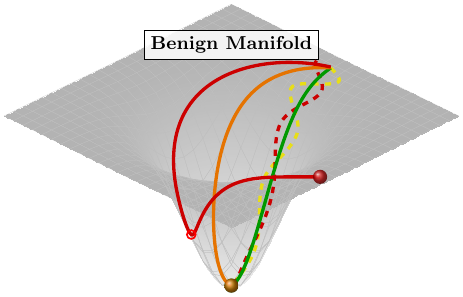}}
    \caption{Illustration of latent representation trajectories relative to the benign manifold. Solid curves denote a normally aligned model, while dashed curves indicate a jailbroken model. Red represents explicit refusal, yellow/orange denotes jailbreak behavior, and green corresponds to benign dialogue.
}
    \label{fig:traj_dynamic}
\end{figure}

\subsection{Stage 2: Benign Manifold and Geometric Turning Angle}
\label{sec:benign_manifold_angle}

To quantify deviations from normal behavior during inference, we introduce geometric metrics. First, we construct a \emph{benign manifold} using representations collected under benign conditions ($B$) as a reference for normal generation (construction details in Appendix~\ref{app:benign_manifold}).

For any given trajectory, we examine two local directions at each step:

\textbf{Trajectory Tangent ($\tau_t^{(A,G)}$):} Reflects the direction of representation evolution during inference.

\textbf{Outward Normal ($n^{(G)}$):} Indicates the direction of maximal deviation from the benign manifold.

We define the \emph{geometric turning angle} as the angle between these vectors:
\begin{equation}
\theta_t^{(A,G)} = \arccos \left(
\frac{\tau_t^{(A,G)\top} n^{(G)}\bigl(h_t^{(A,G)}\bigr)}
{\Vert \tau_t^{(A,G)}\Vert_2 , \Vert n^{(G)}\bigl(h_t^{(A,G)}\bigr)\Vert_2}
\right).
\label{eq:theta}
\end{equation}
Intuitively, $\theta_t^{(A,G)}$ measures whether the trajectory trends toward or away from the benign region. Smaller angles indicate evolution away from the benign manifold, signaling potential safety risk shifts.

\begin{figure}[thbp!]
    \centering
    \includegraphics[width=1\linewidth]{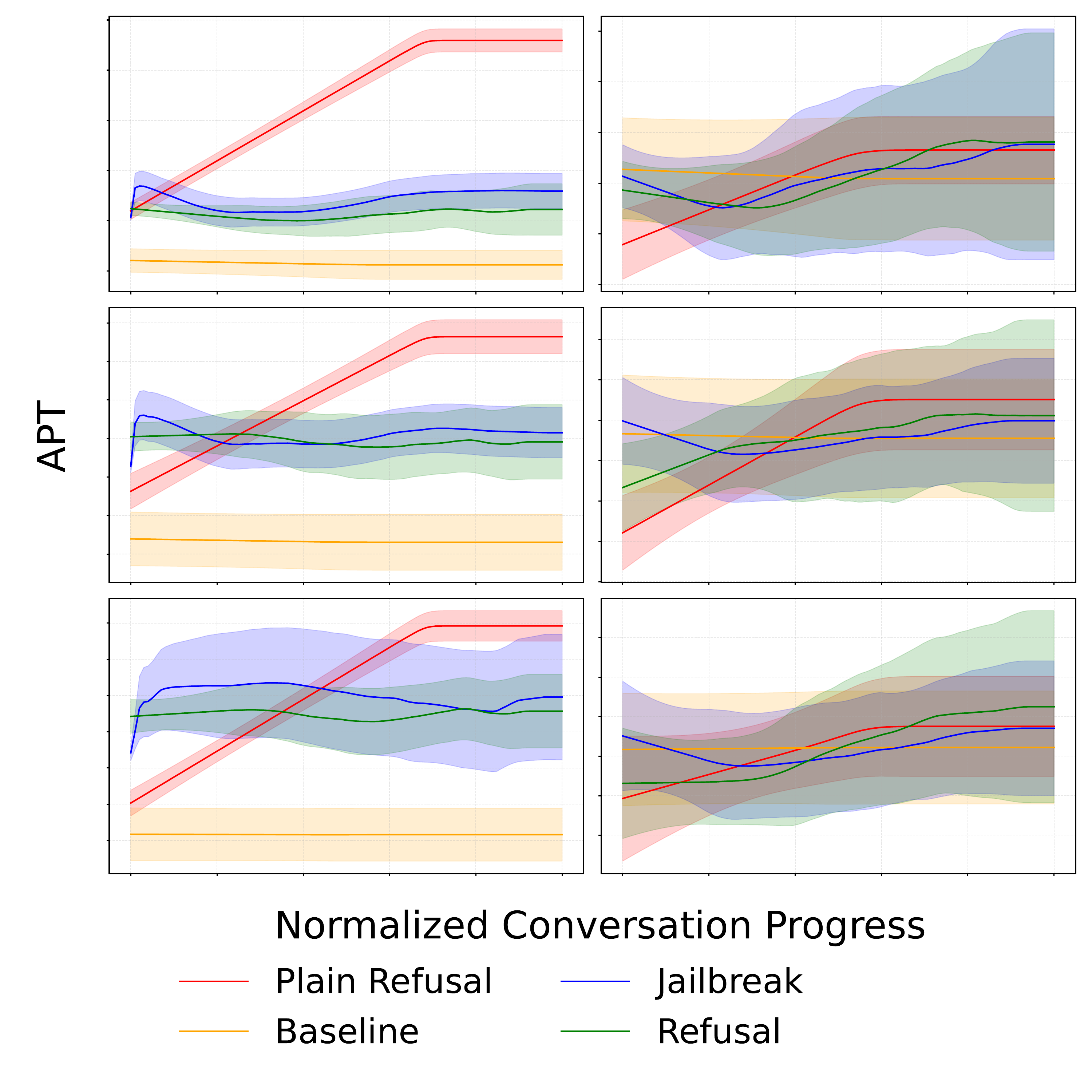}
    \caption{APT trajectories collected from lower/middle/upper(up to down) layer groups. left column correspond to safety-aligned models, while right column ones are from jailbreak-tuned models. Safety-aligned LLMs exhibit well-separated and low-variance APT trajectories, whereas jailbreak-tuned LLMs show weakened separability and substantially higher variance across all query types, indicating degraded safety boundaries. Fig.~\ref{fig:traj_dynamic} further provides an intuitive three-dimensional geometric view of how these trajectories bend and diverge relative to the benign manifold.}
    \label{fig:traj_comb}
\end{figure}

\subsection{Stage 3: Slicing and JSS Metrics Calculation}
\label{sec:probing_conditions}

We quantify safety by analyzing the divergence of turning angle distributions across conditions. We discretize the standardized progress axis $s \in [0,1]$ into $I$ slices $\{s_i\}_{i=1}^I$. For each slice $s_i$, we collect the corresponding angles:

\begin{equation}
\Theta^{(A,G)}(s_i) = {\theta_t^{(A,G)} \mid s_t \in \text{slice } i}.
\label{eq:apt_slice}
\end{equation}

Within each slice $s_i$ and group $G$, we estimate the empirical distributions for two conditions (e.g., $J$ vs. $B$, or $J$ vs. $R$) and compute the Jensen--Shannon divergence (JSD):

\begin{equation}
\label{eq:js_slice}
\mathrm{JS}_i^{(G)}(A,B) = \mathrm{JS}\bigl( \mathrm{D}(\Theta_A^{(G)}(s_i)) ,\big|, \mathrm{D}(\Theta_B^{(G)}(s_i))\bigr).
\end{equation}

To obtain a model-level separability measure, we aggregate JSDs across all slices and layer groups to define the Jensen-Shannon Separability (JSS):

\begin{equation}
\mathrm{JSS}(A,B) = \frac{1}{|G|} \sum_{G} \left( \frac{1}{I} \sum_{i=1}^{I} \mathrm{JS}_i^{(G)}(A,B) \right).
\label{eq:jss_all}
\end{equation}

Algorithm~\ref{alg:jss_core_calc_en} summarizes the core computation for a single model.







\subsection{Insights: Trajectory Separability as a Latent Safety Signal}
\label{sec:latent_safety_interpretation}



\textbf{Separability as an Indicator of Safety Awareness and Refusal Tendency.}
From this geometric perspective, two JSS relations hold key safety implications:

$\mathrm{JSS}(J, B)$ (Jailbreak vs. Benign): Measures whether the model internally distinguishes adversarial inputs from normal dialogue. High separability implies acute latent risk recognition (representation shift details in Appendix~\ref{app:jb_baseline_sep}).

$\mathrm{JSS}(J, R)$ (Jailbreak vs. Ideal Refusal): Reflects the model's intrinsic refusal tendency and resistance to late-stage safety erosion. We observe a significant correlation between this metric and the refusal tendency score (ANM) introduced in Appendix~\ref{app:jb_refusal_sep}.

\textbf{Necessity of Trajectory-Level Aggregation.}
These observations suggest that endpoint or single-step metrics cannot jointly capture both early risk recognition and late refusal stability. Thus, JSS serves as a compact scalar summarizing the latent safety state across the entire inference trajectory, enabling safety evaluation without reliance on text generation.

\section{Experiments and Analysis}
In this section, we employ quantified safety feature metrics to evaluate various aligned models and compare their rankings' correlation and consistency with those obtained from systematic traditional red-teaming evaluations, demonstrating a high degree of consistency in safety scales as well as a significant advantage in evaluation cost. Finally, we investigate the statistical consistency of the ranking results across different datasets and sample sizes, analyzing the stability of our evaluation framework under changes in data distribution and sampling conditions.

\begin{figure*}
    \centering
    \scalebox{0.75}{\includegraphics[width=1\linewidth]{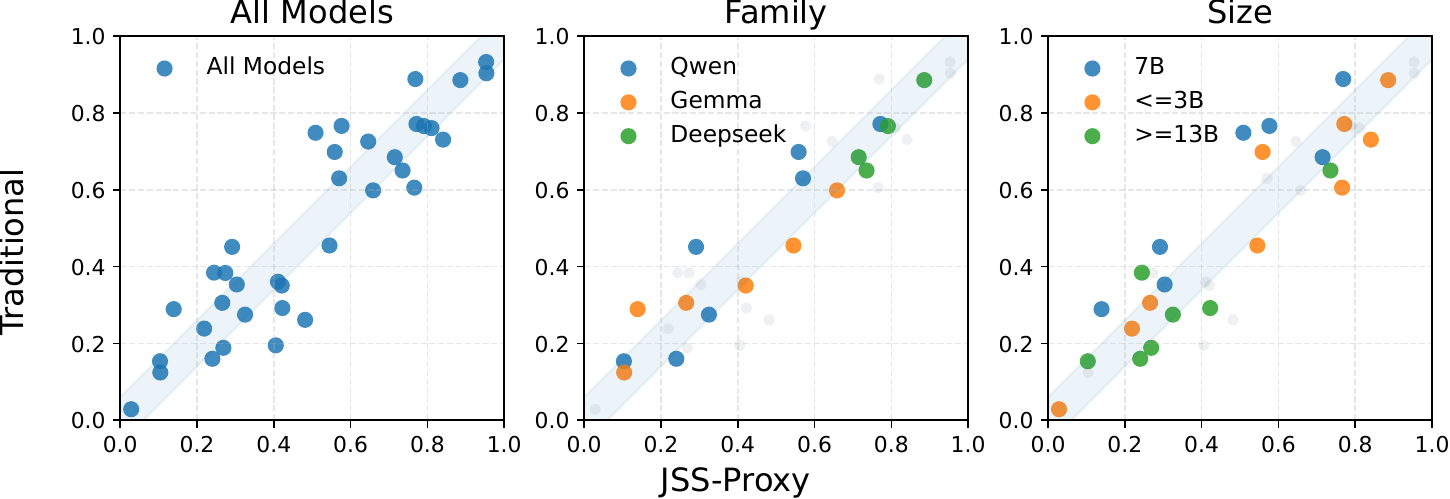}}
    \caption{Multi-perspective cross-model safety ranking consistency between N-GLARE and traditional red-teaming. Detailed figures and datapoints at Appendix Fig.~\ref{fig:combined_ranking_from_diff_LLMfamilies_appendix} and Appendix Table~\ref{tab:jb_safety_benchmark_split_A} ~\ref{tab:jb_safety_benchmark_split_B}.}
    \label{fig:combined_ranking_from_diff_LLMfamilies}
\end{figure*}

\subsection{Model Differentiation and Safety Characterization}
This section presents the core capability of our evaluation framework: its ability to precisely and transferably characterize safety differences across models. Within this overarching framework, our analyses proceed along two complementary axes: Inter-Model Dimension (Model-Level Differentiation): examining variations in latent-space separability among different kinds of models to reveal the rank correlation between our method and large-scale traditional red-teaming trends; Intra-Model Dimension (DPO Fine-Tuning Dynamics): analyzing how changes induced by DPO-based fine-tuning are reflected in our JSS proxy metric, thereby testing the sensitivity and faithfulness of JSS to alignment-related weight updates.

\begin{figure}[htbp]
    \centering
    \includegraphics[width=1\linewidth]{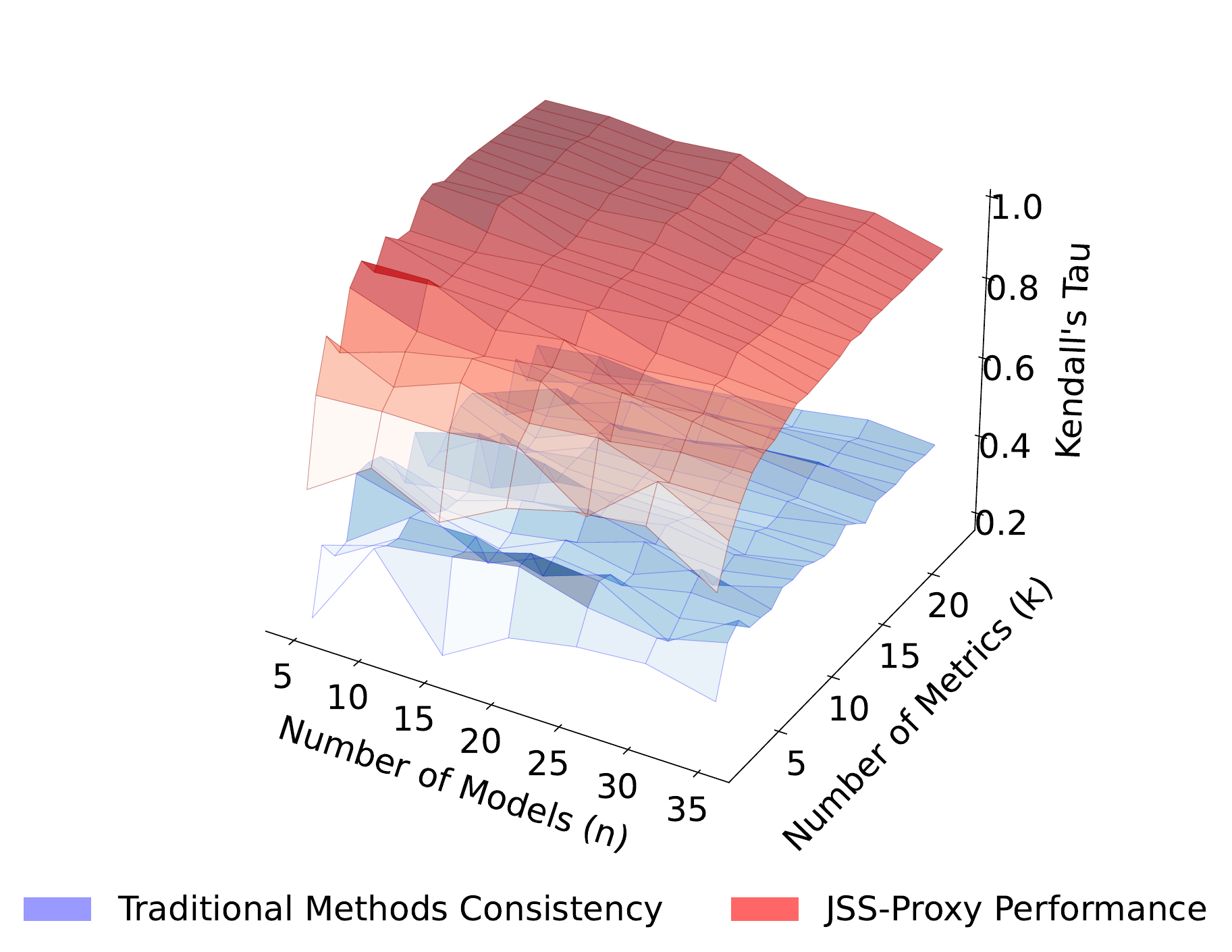}
    \caption{Comparison of ranking consistency between JSS-proxy and traditional red-teaming metrics across varying numbers of models ($n$) and evaluation metrics ($k$), measured by Kendall’s $\tau$.}
    \label{fig:jss_vs_traditional_tau_consistency}
\end{figure}

\subsubsection{Cross-Model Validation of JSS-Derived Safety Proxies}

For model-level differentiation, we collected a diverse set of models and evaluated their performance under a series of standardized red-teaming safety tests (20+ testing protocols with over 7000 test cases, more data listed in the Appendix table ~\ref{tab:jb_safety_split_A} and ~\ref{tab:jb_safety_split_B}). For each model, the results were aggregated into a single averaged traditional safety score. Our goal in this experiment is twofold: (1) to derive proxy safety indicators from our JSS-based evaluation framework, and (2) to assess the consistency between the proxy-based ranking and the ranking obtained from conventional red-teaming safety evaluations.

Building upon the JSS evaluation indices defined earlier, we construct two higher-level proxy metrics that capture complementary aspects of safety separability.
\begin{equation}
\mathrm{JB/PB\ Ratio}
=
\frac{\sum_{s\in[0,1]}\ \mathrm{JSS}_s\big(\mathrm{J},\mathrm{B}\big)}
     {\sum_{s\in[0,1]}\ \mathrm{JSS}_s\big(\mathrm{P},\mathrm{B}\big)}
\end{equation}

A higher value of $\mathrm{JB/PB\ Ratio}$ indicates that Jailbreak trajectories are more strongly separated from the Baseline than PlainQuery trajectories, reflecting the model’s latent ability to discriminate adversarial jailbreak behavior beyond explicit harmful intent.

\begin{equation}
\mathrm{JR\ Min/Max}
=
\frac{\min_{s\in[0,1]}\ \mathrm{JSS}_s\big(\mathrm{J},\mathrm{R}\big)}
     {\max_{s\in[0,1]}\ \mathrm{JSS}_s\big(\mathrm{J},\mathrm{R}\big)}
\end{equation}

A higher value of $\mathrm{JR\ Min/Max}$ indicates that Jailbreak–Ideal Refusal separability remains stable across the entire trajectory, whereas a low value reflects late-stage collapse of refusal and successful jailbreak intrusion.

We quantify the agreement between our proxy-based safety ranking $r_{\text{jss-proxy}}$ and the ranking obtained from conventional red-teaming evaluations $r_{\text{Hybrid}}$ using Kendall’s $\tau$ correlation coefficient:
$\tau(r_{\text{jss-proxy}}, r_{\text{Hybrid}}) \in [-1,1].$

Figure ~\ref{fig:combined_ranking_from_diff_LLMfamilies} show that cross a sufficiently large model set, the proxy-based ranking exhibits strong concordance with the red-teaming ground truth. This demonstrates that our JSS-derived proxy indicators can accurately reproduce traditional safety orderings within representative alignment regimes, while requiring far lower evaluation cost. 

As the model and metric pool expands in Figure ~\ref{fig:jss_vs_traditional_tau_consistency}, $\tau$ remains consistently high, with $p$-values far below conventional significance thresholds($< 0.05$). This indicates that even under diverse models and red-teaming metrics, the consistency between the proxy-based and red-teaming rankings remains higher than red-teaming metrics themselves and statistically significant.

\begin{figure*}
    \centering
    \includegraphics[width=1\linewidth]{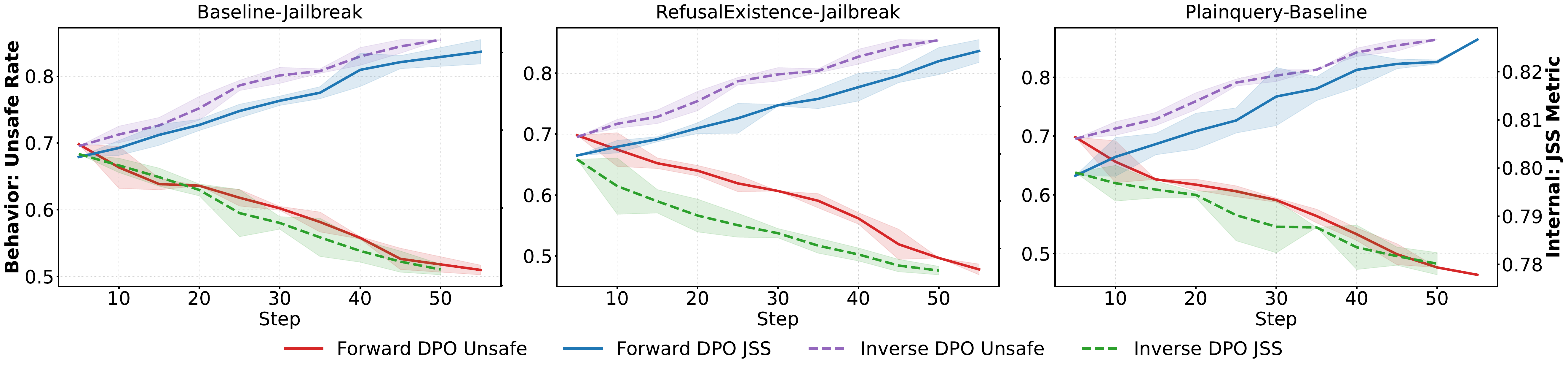}
    \caption{\textbf{Evolution of JSS and behavioral safety metrics under Forward vs. Inverse DPO.} The dynamic evolution of both surface-level safety behavior metrics (Unsafe Rate) and latent proxy metrics (JSS). The three columns respectively correspond to separability relations across different dialogue families. This confirms that JSS reliably tracks the direction of safety strengthening or weakening, providing a stable latent–surface correspondence.}
    \label{fig:DPO-JSS evolution matrix}
\end{figure*}

\subsubsection{Consistency under Model Weight Variations}
In essence, we demonstrate that the geometric safety scale captured by JSS not only reflects static differences among models but also dynamically responds to behavioral transitions induced by weight adjustments. 
We introduce two opposite preference optimization directions—Forward DPO and Inverse DPO—to respectively enhance and attenuate safety alignment by fine-tuning the same base model with reversed preferences. Both settings use the standard DPO objective:
$\mathcal{L}_{\mathrm{DPO}}(\theta)
= -\,\mathbb{E}_{(x,y^+,y^-)\sim\mathcal{D}}
\Big[
  \log\sigma\!\big(\beta\,\Delta_\theta\big)
\Big]$
$\text{where }\;
\Delta_\theta
= \log p_\theta(y^+\!\mid\!x)
  - \log p_\theta(y^-\!\mid\!x)$
In Forward DPO, safety-aligned responses are preferred over harmful ones ($y^+$ = safe, $y^-$ = harmful), whereas Inverse DPO reverses this preference to favor harmful responses ($y^+$ = harmful, $y^-$ = safe).

To avoid conclusions driven by a single fine-tuning run or a single training step, we analyze the training-time co-evolution between a latent proxy (JSS) and surface safety metrics (UR and RR) under both Forward and Inverse DPO.

Figure~\ref{fig:DPO-JSS evolution matrix} shows that JSS changes track the whole training dynamics of safety behavior: when safety improves under Forward DPO, JSS increases together with decreased unsafe surface metrics; when safety is weakened under Inverse DPO, JSS decreases as surface safety degrades (Unsafe Rate increases). Importantly, the turning points of the dashed JSS curves appear earlier than the corresponding shifts in output behavior-based metrics (detailed information in the Appendix Table~\ref{tab:jss_lag_analysis}). Figure~\ref{fig:dpo-jss-heapmap-matrix} further supports this temporal relation at the step level: JSS exhibits strong step-wise coupling with both UR and RR across layers (Unsafe Rate and Refusal Rate definition listed in Appendix~\ref{appendix sec:UR-RR-Def}).

Together, these results indicate that latent separability (JSS) changes earlier than refusal-rate changes during training, making JSS a practical early-warning signal for upcoming safety degradation, while unsafe-rate changes largely reflect the same-step training state.

\begin{figure}[htbp]
    \centering
    \includegraphics[width=0.75\linewidth]{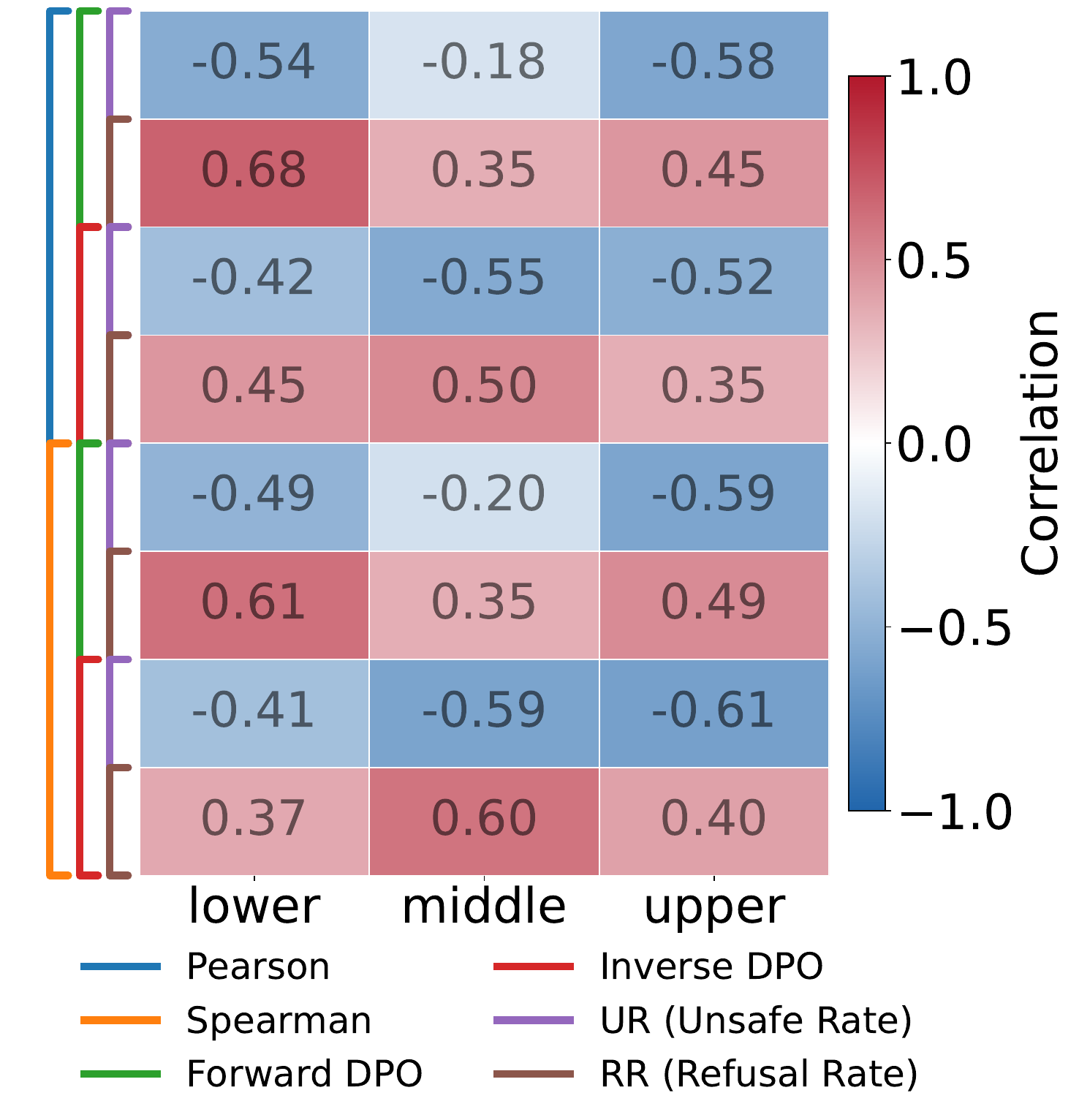}
    \caption{Step-Wise Safety Metrics (UR: Unsafe Rate; RR: Refusual Rate) and JSS Correlation Heatmap. JSS exhibits strong step-wise coupling with both UR and RR across layers}
    \label{fig:dpo-jss-heapmap-matrix}
\end{figure}
















\begin{table}[t]
\centering
\resizebox{\linewidth}{!}{
\begin{tabular}{lccc}
\toprule
Attack &
$\mathrm{tok}_{\mathrm{in}}$ (k) &
$\mathrm{tok}_{\mathrm{out}}$ (k) &
$\mathrm{tok}$ (k) \\
\midrule

ArtPrompt
 & 50.92\scriptsize (159)
 & 70.10\scriptsize (216)
 & 121.02\scriptsize (375) \\

Cipher
 & 515.65\scriptsize (--)
 & 151.90\scriptsize (63)
 & 667.55\scriptsize (63) \\

DAN
 & 502.69\scriptsize (--)
 & 99.76\scriptsize (59)
 & 602.45\scriptsize (59) \\

DevMode
 & 519.71\scriptsize (--)
 & 59.74\scriptsize (48)
 & 579.45\scriptsize (48) \\

MultiLing
 & --\scriptsize (--)
 & 167.87\scriptsize (17)
 & 167.87\scriptsize (17) \\

\midrule

AutoDAN
 & 1354.89\scriptsize (129)
 & 2224.48\scriptsize (185)
 & 3579.37\scriptsize (314) \\

GPTFuzzer
 & 836.48\scriptsize (82)
 & 700.46\scriptsize (76)
 & 1536.94\scriptsize (158) \\

PAIR
 & 1212.90\scriptsize (123)
 & 775.71\scriptsize (85)
 & 1988.61\scriptsize (207) \\

\midrule

Mixed (PyRIT)
 & --
 & --
 & 4797.30\scriptsize (--) \\

\midrule

Ours (JSS-only)
 & \textbf{--}\scriptsize (--)
 & \textbf{1.03}\scriptsize (--)
 & \textbf{1.03}\scriptsize (--) \\

\bottomrule
\end{tabular}
}

\caption{Attack statistics aggregated over single vs.\ multi-round settings.
Each entry is reported in thousands of tokens (k{\scriptsize (avg)}), where
$\mathrm{tok}_{\mathrm{in}}$ denotes input tokens to the subject model,
$\mathrm{tok}_{\mathrm{out}}$ denotes output tokens generated for evaluation,
and $\mathrm{tok}$ denotes their sum.}
\label{tab:attacks-aggregated}
\end{table}

\subsection{Baseline Comparison: Cost and Efficiency of Traditional Red-Teaming vs. Our JSS-Only Pipeline}

We compare the evaluation cost of our JSS-only pipeline with traditional online red-teaming under matched task scale and sample size.

The baseline methods that contain single-turn, multi-turn and mixed strategy follow a standard red-teaming workflow with separate attacker, subject, and evaluator models, while our method relies solely on a several forward passes of the subject model on offline attack trajectories, without online generation or auxiliary evaluator inference.
Specifically, we provide more details of the cost model, token accounting, and timing formulation in Appendix~\ref{app:cost_model} when we conduct the experiment on mixed strategy (PyRIT~\cite{munoz2024pyritframeworksecurityrisk}). The results support our method's advantages as well in Appendix Figure~\ref{fig:pyrit_cost_analyze}.

In table ~\ref{tab:attacks-aggregated} we can see that across all red-teaming intensities and evaluation settings, the JSS-only pipeline consistently achieves orders-of-magnitude reductions in both token consumption.
This efficiency gain stems from eliminating iterative generation and multi-model interaction, thereby removing the primary scalability bottleneck of traditional red-teaming.

\begin{figure}[t]
    \centering
    \includegraphics[width=1\linewidth]{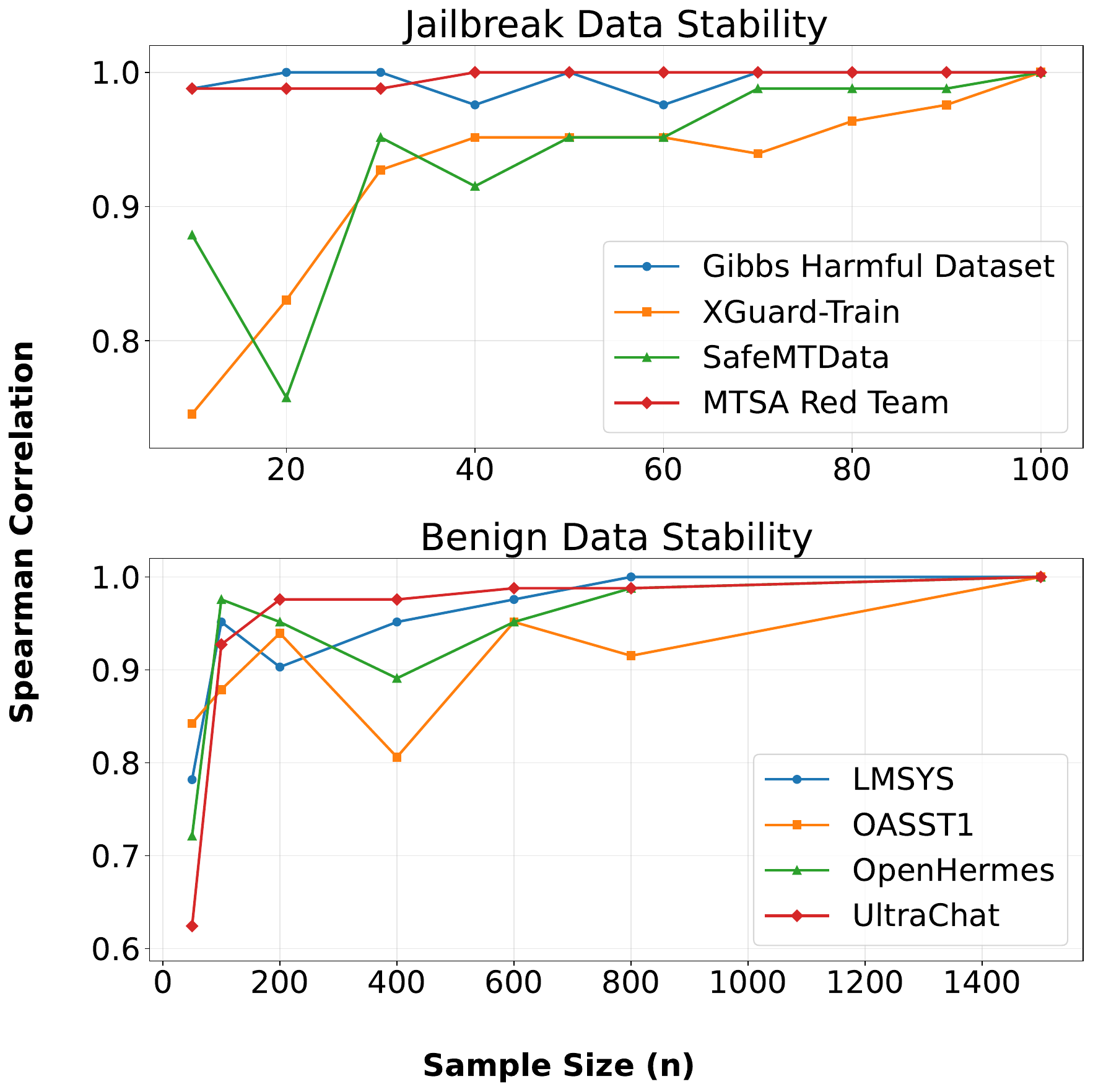}
    \caption{Rank correlation stability of JSS-based safety ordering under varying sample sizes for jailbreak and benign data. Appendix Fig~\ref{fig:2x2matrixDataset_only_appendix} has more information about this.}
    \label{fig:rank_corr_with_n}
\end{figure}

\subsection{Model Sensitivity and Systemic Stability Experiments}

If our method is indeed capturing an intrinsic property of the model’s alignment dynamics—rather than dataset-specific artifacts—then its induced safety ordering should remain stable when we perturb the underlying data and sampling conditions. The goal of this section is therefore to test the \emph{systemic stability} of our JSS-based safety scale. Concretely, we ask:
(i) whether the relative JSS differences between models remain consistent when we vary which jailbreak trajectories are used under different attack datasets and how many sample size $n$ trajectories are included; and
(ii) whether the same stability holds when we perturb the benign reference distribution, by changing the source corpus and the number of benign samples used to estimate the natural manifold.

As shown in Fig~\ref{fig:rank_corr_with_n} and Table~\ref{tab:rank_corr_total_final}, rank correlations consistently increase with sample size and rapidly converge to high values, indicating that the induced model ordering is not sensitive to the specific choice of attack or benign datasets.

Lower correlations mainly arise under heterogeneous datasets or small sample sizes—particularly for jailbreak data—where weak or localized separability amplifies noise, but these effects vanish as rankings converge at moderate to large $n$.

\begin{table}[t]
    \centering

    \resizebox{\linewidth}{!}{
    \begin{tabular}{ccccccc}
        \toprule
        Source & Metric & Matrix Dim & Mean & Std & Q25 & Q75 \\
        \midrule
        \multirow{6}{*}{benign} & \multirow{3}{*}{kendall} 
         & n & 0.78 & 0.11 & 0.65 & 0.87 \\
         &  & dataset & 0.67 & 0.14 & 0.56 & 0.78 \\
         &  & model & 0.73 & 0.36 & 0.33 & 1.00 \\
         & \multirow{3}{*}{spearman} 
         & n & 0.88 & 0.08 & 0.79 & 0.94 \\
         &  & dataset & 0.79 & 0.11 & 0.72 & 0.87 \\
         &  & model & 0.76 & 0.34 & 0.40 & 1.00 \\
        \midrule
        \multirow{6}{*}{jailbreak} & \multirow{3}{*}{kendall} & n & 0.87 & 0.14 & 0.78 & 0.96 \\
         &  & dataset & 0.45 & 0.24 & 0.33 & 0.60 \\
         &  & model & 0.54 & 0.12 & 0.47 & 0.61 \\
         & \multirow{3}{*}{spearman} & n & 0.94 & 0.08 & 0.92 & 0.99 \\
         &  & dataset & 0.57 & 0.28 & 0.42 & 0.79 \\
         &  & model & 0.68 & 0.12 & 0.58 & 0.78 \\
        \bottomrule
    \end{tabular}}

    \caption{Summary of rank correlation statistics for JSS-based safety ordering.
The \textit{Matrix Dim} column indicates the axis along which rankings are compared:
\emph{n} denotes varying sample sizes,
\emph{dataset} denotes different jailbreak or benign data sources,
and \emph{model} denotes comparisons across model variants.
Statistics are reported as mean, standard deviation, and interquartile range (Q25–Q75). A detailed pair-wise heatmap is at Appendix Fig~\ref{fig:sensitivity_heatmap}.}
    \label{tab:rank_corr_total_final}
\end{table}

\section{Conclusion}
This paper presents N-GLARE, a non-generative, representation-based framework for LLM safety evaluation that captures safety state through latent trajectory geometry rather than outputs. By modeling a benign manifold and using JSS to measure deviations of jailbreak and refusal trajectories, N-GLARE produces model-level safety rankings consistent with large-scale red-teaming across over 40 models, while reducing cost by more than two orders of magnitude. It further reveals fine-grained safety dynamics, demonstrating latent trajectory analysis as a scalable tool for continuous safety diagnostics.

\section{Limitations}
While N-GLARE demonstrates strong empirical effectiveness, several limitations remain. First, although our evaluation covers over 40 models with diverse sizes and alignment settings, it does not exhaustively represent all architectures, training paradigms, or deployment variants. Second, N-GLARE requires access to internal hidden representations, which limits its direct applicability to fully black-box models that expose only text outputs. In addition, the geometric abstraction underlying N-GLARE—modeling benign behavior as a low-dimensional manifold and inferring safety from trajectory separability—may oversimplify more complex latent structures in large models. Finally, although the framework avoids output generation, it still depends on the design of probing conditions, whose coverage and quality can influence the resulting safety indicators.

The limitations above also point to several promising directions for future research. Model coverage can be naturally expanded as new architectures and checkpoints become available, enabling broader validation of representation-based safety indicators. The reliance on hidden states motivates future work on approximate or surrogate signals that could extend N-GLARE–style analysis to partially observable or API-only settings. Beyond harmfulness and refusal, the trajectory-based framework can be generalized to capture other alignment dimensions, such as bias, factual reliability, or long-horizon agentic safety, by defining task-specific probing conditions and reference manifolds. More expressive geometric models, including non-linear or multi-manifold representations, may further improve fidelity to complex latent structures. Finally, systematic and automated probing design offers a new research direction, positioning latent trajectory analysis as a foundation for scalable, continuous alignment monitoring and comparative safety evaluation.

\bibliography{custom}

\appendix




\twocolumn

\section{Unsafe Rate and Refusual Rate Definition}
\label{appendix sec:UR-RR-Def}
\textbf{Unsafe Rate.} This metric is computed using the safety classifications provided by \texttt{Qwen3Guard}. 
A model output is counted as unsafe whenever the guard model labels it as ``Unsafe.'' Formally,
\[
\text{Unsafe Rate} = \frac{N_{\text{Unsafe}}}{N_{\text{Total}}},
\]
where \(N_{\text{Unsafe}}\) denotes the number of responses flagged as Unsafe by the guard model, and 
\(N_{\text{Total}}\) denotes the total number of model responses evaluated.

\textbf{Refusal Rate.} This metric is based on refusal-style keyword matching. 
A model response is counted as a refusal if it contains typical refusal patterns such as 
``I'm sorry'', ``I cannot comply'', ``I refuse to answer'', or similar formulations. Formally,
\[
\text{Refusal Rate} = \frac{N_{\text{Refusal}}}{N_{\text{Total}}},
\]
where \(N_{\text{Refusal}}\) denotes the number of responses that match the refusal keyword patterns.

\textbf{Complementarity.} These two metrics capture opposite behavioral tendencies and are therefore complementary. 
Lower \emph{Unsafe Rate} indicates stronger safety, while lower \emph{Refusal Rate} indicates better helpfulness and less over-defensiveness. 
A high Unsafe Rate signals safety failures, whereas a high Refusal Rate reflects an overly conservative model that rejects too many user requests.

\section{Empirical Analysis of Latent Safety Trajectories}
\label{app:latent_safety_analysis}

This appendix provides structured empirical evidence and metric definitions supporting the methodological insights presented in Section~\ref{sec:latent_safety_interpretation}. All observations are validated through quantitative analysis on latent representation trajectories.

\subsection{Trajectory Evolution Across Dialogue Families}
\label{app:traj_evolution}

\paragraph{Observation.}
Model safety manifests as a dynamic process in latent representation space rather than a static output attribute.

\paragraph{Protocol.}
We extract APT trajectories for four dialogue families:
\emph{Baseline}, \emph{PlainQuery}, \emph{Jailbreak}, and \emph{Ideal Refusal}.
Each trajectory is traced across layers and decoding steps under identical model settings.

\paragraph{Empirical Evidence.}
Figure~\ref{fig:traj_comb} shows that:
\begin{itemize}
    \item Baseline trajectories remain flat and low-variance, forming a stable reference manifold.
    \item PlainQuery trajectories diverge sharply from Baseline across all layers, indicating rapid risk detection and refusal activation.
    \item Jailbreak trajectories gradually deviate from Baseline and exhibit downward drift in higher layers, reflecting progressive erosion of refusal mechanisms.
    \item Ideal Refusal initially overlaps with Jailbreak but later bifurcates, indicating a latent distinction between compliant and defensive generation paths.
\end{itemize}

\paragraph{Takeaway.}
Safety-relevant information is encoded in the relative geometry of trajectories rather than in isolated representations.

\subsection{Jailbreak--Baseline Separability and Detection Capability}
\label{app:jb_baseline_sep}

\paragraph{Observation.}
Separability between Jailbreak and Baseline trajectories reflects the model’s latent capability to detect jailbreak process shifting.

\paragraph{Metric and Protocol.}
To test whether Jailbreak trajectories encode stage-dependent information, we partition the jailbreak process into \emph{early}, \emph{mid}, and \emph{late} stages.
A lightweight neural classifier is trained to distinguish representations from different stages.
If stage-wise structure is present, classification accuracy should exceed chance level.

\paragraph{Empirical Evidence.}
For aligned LLaMA-family models, stage classification accuracy reaches:

\begin{table}[t]
\centering
\caption{Stage-wise classification accuracy for Jailbreak trajectory representations in aligned LLaMA-family models.}
\label{tab:jb_stage_classification}
\begin{tabular}{l c}
\hline
\textbf{Stage Pair} & \textbf{Classification Accuracy} \\
\hline
Early vs.\ Mid  & 0.86 -- 0.97 \\
Early vs.\ Late & 0.91 -- 0.96 \\
Mid vs.\ Late   & 0.85 -- 0.86 \\
\hline
\end{tabular}
\end{table}
\noindent
Higher-than-chance accuracy across all stage pairs indicates a reliable and alignment-dependent phase structure in latent jailbreak trajectories.

After ablating safety alignment (manually jailbroken variants), accuracy collapses to approximately $0.5$ for all pairs.

\paragraph{Takeaway.}
Jailbreak trajectories exhibit a reliable, alignment-dependent phase structure, supporting the use of Jailbreak--Baseline separability as a proxy for latent detection capability.

\subsection{Jailbreak--Ideal Refusal Separability and Refusal Tendency}
\label{app:jb_refusal_sep}

\paragraph{Observation.}
Separability between Jailbreak and Ideal Refusal trajectories reflects intrinsic refusal tendency.

\paragraph{Metric Definition (ANM).}
We quantify refusal inclination using the \emph{Activation-based Negation Measure (ANM)}:
\begin{equation}
\mathrm{ANM}(x)
= \sum_{v\in\mathcal V} W_{\mathrm{ref}}(v)\, p(v\mid x),
\end{equation}
where \(p(v\mid x)\) is the next-token distribution and \(W_{\mathrm{ref}}\) is a fixed refusal-prototype distribution constructed from normalized embeddings of refusal seed phrases.

\begin{figure}[htbp]
    \centering
    \includegraphics[width=1\linewidth]{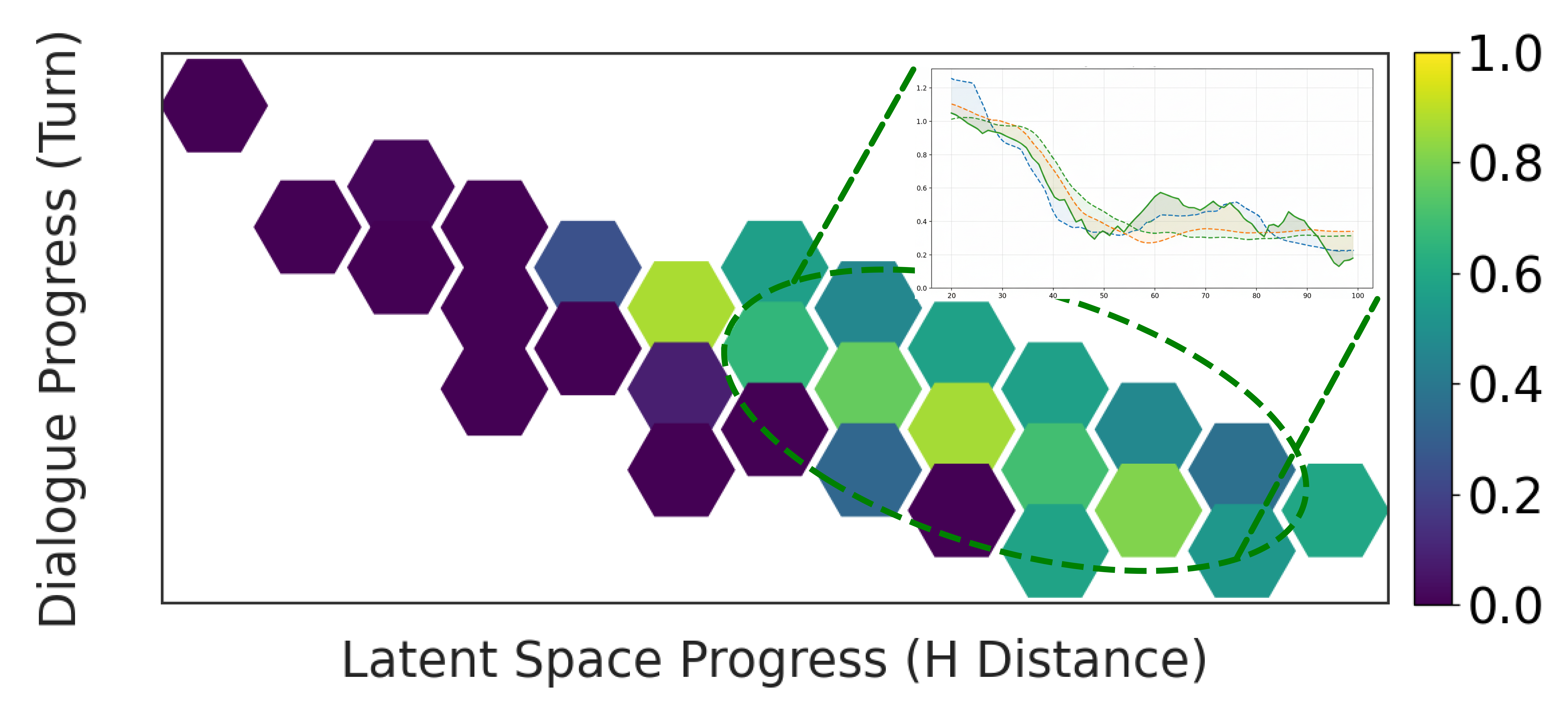}
    \caption{\textbf{Coupled dynamics of JSS and ANM along the jailbreak process.} The color indicates the normalized refusal-intensity score (ANM); The inset line plot in the upper-right overlays slice-wise JSS and ANM curves over the standardized progress axis, revealing a shared mid-phase rise and late-phase collapse, which highlights the strong correlation between geometric separability (JSS) and probability-level refusal tendency (ANM).}
    \label{fig:jss-anm-relation}
\end{figure}

\paragraph{Empirical Evidence.}
Across layers, ANM curves exhibit a consistent mid-stage rise followed by late-stage collapse.
This pattern aligns with the Ideal Refusal--Jailbreak JSS waveform.
Layer-wise correlations between ANM and Ideal Refusal--Jailbreak JSS are:
\[
r = 0.70,\ 0.56,\ 0.54 \quad (\text{RMSE} = 0.13\text{--}0.20).
\]
Figure ~\ref{fig:jss-anm-relation} also shows that mid-stage elevation corresponds to increased refusal probability, while late-stage decline coincides with jailbreak success.

\paragraph{Takeaway.}
Latent trajectory separability is tightly coupled with explicit refusal behavior, validating its use as an interpretable proxy for refusal stability.

\section{Construction of the Benign Manifold and Geometric Turning Angle}
\label{app:benign_manifold}

This appendix provides the full mathematical formulation underlying the benign
manifold, the associated normal direction, and the geometric turning angle
introduced in Section~\ref{sec:benign_manifold_angle}. These derivations are
omitted from the main text for clarity and readability.

\subsection{Benign Manifold Construction}
\label{app:benign_manifold_construction}

To characterize normal generation behavior, we construct a benign manifold using
latent representations collected under benign inputs.
For each group $G$, we aggregate all benign activations across prompts and
progress values:
\begin{equation}
\mathcal{H}_B^{(G)} = \{h^{(B,G)}\}.
\end{equation}

We estimate the empirical mean and covariance:
\begin{equation}
\begin{split}
\mu^{(G)} &= \mathbb{E}[h^{(B,G)}], \\
\Sigma^{(G)} &= \mathbb{E}\bigl[(h^{(B,G)}-\mu^{(G)})(h^{(B,G)}-\mu^{(G)})^\top\bigr].
\end{split}
\end{equation}

A rank-$r$ eigendecomposition is then performed:
\begin{equation}
\Sigma^{(G)} \approx U_r^{(G)} \Lambda_r^{(G)} {U_r^{(G)}}^\top,
\qquad r \ll d,
\end{equation}
where $U_r^{(G)}$ contains the top-$r$ eigenvectors and $\Lambda_r^{(G)}$ the
corresponding eigenvalues.
This defines a linear surrogate of the benign manifold:
\begin{equation}
\mathcal{S}^{(G)} = \mathrm{span}\bigl(U_r^{(G)}\bigr).
\end{equation}

By the Eckart--Young--Mirsky theorem~\cite{EckartYoung1936,GolubVanLoan2013},
this subspace provides the optimal rank-$r$ approximation of the benign
distribution in the least-squares sense.
Under standard smoothness assumptions, it also serves as a local tangent-space
approximation of the underlying benign manifold~\cite{DoCarmo1976}.

\subsection{Residual Energy and Outward Normal}
\label{app:residual_normal}

Given a representation $h$, its reconstruction from the benign subspace is:
\begin{equation}
\hat{h}^{(G)} = \mu^{(G)} + U_r^{(G)} {U_r^{(G)}}^\top (h - \mu^{(G)}).
\end{equation}

The residual vector
\begin{equation}
r^{(G)}(h) = h - \hat{h}^{(G)}
\end{equation}
captures the component orthogonal to the benign manifold.
We define the residual energy:
\begin{equation}
E_b^{(G)}(h) = \Vert r^{(G)}(h) \Vert_2^2.
\end{equation}

The gradient of this energy with respect to $h$ is:
\begin{equation}
n^{(G)}(h) = \nabla E_b^{(G)}(h) = 2\, r^{(G)}(h),
\end{equation}
which defines an outward normal direction pointing away from the benign manifold.
Up to a constant factor, the residual vector itself therefore serves as the
normal direction used in practice.

\subsection{Trajectory Tangent Direction}
\label{app:trajectory_tangent}

For a representation trajectory
$\Gamma_M^{(A,G)} = \{h_t^{(A,G)}\}_{t=1}^{T_A}$,
the local tangent direction at step $t$ is defined as:
\begin{equation}
\tau_t^{(A,G)} =
\frac{h_{t+1}^{(A,G)} - h_t^{(A,G)}}
     {\Vert h_{t+1}^{(A,G)} - h_t^{(A,G)} \Vert_2},
\qquad 1 \le t < T_A.
\end{equation}

This tangent captures the instantaneous direction of representation evolution
along the inference process.

\subsection{Derivation of the Geometric Turning Angle}
\label{app:turning_angle_derivation}

At each step $t$, we measure how the trajectory direction aligns with the outward
normal of the benign manifold.
The geometric turning angle is defined as:
\begin{equation}
\theta_t^{(A,G)} =
\arccos \left(
\frac{\tau_t^{(A,G)\top} n^{(G)}\bigl(h_t^{(A,G)}\bigr)}
     {\Vert \tau_t^{(A,G)}\Vert_2 \,
      \Vert n^{(G)}\bigl(h_t^{(A,G)}\bigr)\Vert_2}
\right).
\end{equation}

Since $n^{(G)}(h)$ is collinear with $r^{(G)}(h)$, the angle can equivalently be
computed using the residual direction.
By construction, $\theta_t^{(A,G)} \in [0,\pi]$.
Smaller angles indicate that the trajectory is moving more directly away from
the benign manifold, corresponding to stronger deviation from normal generation
behavior.

This turning angle serves as the local geometric primitive for trajectory-level
aggregation in the main text.

\section{Cost Model and Efficiency Analysis}
\label{app:cost_model}

This appendix details the cost modeling assumptions, token accounting rules, and
runtime decomposition used in the efficiency comparison between the traditional
online red-teaming pipeline and our JSS-only evaluation pipeline.

All measurements are conducted under matched task scale and sample size to ensure
a fair comparison.

\subsection{Evaluation Pipelines and Model Roles}
\label{app:cost_pipeline}

We compare two evaluation pipelines:

\paragraph{JSS-only pipeline (Ours).}
Our method operates on offline multi-turn attack trajectories.
For each example, we perform a single forward pass on the subject model $M_s$
to extract hidden representations.
No online generation, attack synthesis, or evaluator inference is required.

\paragraph{Traditional red-teaming pipeline (Baseline).}
The baseline follows a standard online red-teaming workflow implemented using
PyRIT with Ollama backends, involving three model roles:
\begin{itemize}
    \item $M_r$: red-team attack generator,
    \item $M_s$: subject model under evaluation,
    \item $M_e$: evaluator or safety discriminator.
\end{itemize}
Each attack session proceeds in multiple turns, with all three models invoked
at every turn.

\subsection{Token Cost Formulation}
\label{app:cost_token}

Let $N$ denote the number of attack targets, $\bar{T}_r$ the average number of
attack turns per session, and $\bar{L}_r$, $\bar{L}_s$, $\bar{L}_e$ the average
token lengths generated by $M_r$, $M_s$, and $M_e$, respectively.

\paragraph{Baseline.}
The total generated-token cost of traditional red-teaming is:
\begin{equation}
C_{\text{token}}^{\text{Base}}(N)
= N \cdot \bar{T}_r \cdot
\bigl(\bar{L}_r + \bar{L}_s + \bar{L}_e \bigr).
\end{equation}

\paragraph{JSS-only pipeline.}
Our method requires only a forward pass over an offline multi-turn dataset.
For bookkeeping consistency, we count one subject-model forward pass per turn
as one token-equivalent unit:
\begin{equation}
C_{\text{token}}^{\text{Ours}}(N)
= N \cdot \bar{T}_{\text{offline}},
\end{equation}
where $\bar{T}_{\text{offline}}$ denotes the average number of turns in the
offline attack trajectories.

\subsection{Runtime Cost Decomposition}
\label{app:cost_time}

\paragraph{Baseline.}
The total runtime cost of the traditional pipeline decomposes as:
\begin{equation}
C_{\text{time}}^{\text{Base}}
= \mathrm{Time}_{M_r}
+ \mathrm{Time}_{M_s}
+ \mathrm{Time}_{M_e},
\end{equation}
where each term includes both model inference time and associated system
overheads incurred at every attack turn.

\paragraph{JSS-only pipeline.}
The runtime cost of our method is:
\begin{equation}
C_{\text{time}}^{\text{Ours}}
= \mathrm{Time}_{M_s^{\text{forward}}}
+ \mathrm{Time}_{\text{offline\_JSS}}
+ \mathrm{Time}_{\text{benign\_con}}.
\end{equation}
Here, $\mathrm{Time}_{M_s^{\text{forward}}}$ denotes the time required to perform
forward passes on the subject model over the offline dataset.
The costs of benign distribution construction and JSS computation are typically
negligible compared to model inference and are therefore omitted in efficiency
comparisons.

\subsection{Measurement Protocol}
\label{app:cost_protocol}

We evaluate both pipelines at multiple red-teaming intensities
$N \in \{n_1, n_2, \dots\}$.

For the baseline, online attacks are generated using PyRIT, and all generated
tokens from $M_r$, $M_s$, and $M_e$ are counted.
Wall-clock time is measured end-to-end for each evaluation run.

For the JSS-only pipeline, the same set of attack instructions is used in the
form of offline multi-turn datasets derived from RefEx and PlainR.
We record token-equivalent counts and elapsed wall-clock time for the forward
pass and offline analysis.

\subsection{Evaluation Metrics}
\label{app:cost_metrics}

We compare the two pipelines using the following efficiency metrics:
\begin{itemize}
    \item \textbf{Total generated-token count:} overall system token usage.
    \item \textbf{Subject-model token share:} fraction of tokens attributed to
    the evaluated model $M_s$.
    \item \textbf{Per-token time cost:} wall-clock time normalized by total
    generated tokens, serving as a throughput indicator.
\end{itemize}

Lower per-token time cost indicates higher evaluation efficiency.

\subsection{Summary of Efficiency Results}
\label{app:cost_summary}

Across all red-teaming intensities and evaluation settings, the JSS-only pipeline
consistently reduces both token consumption and runtime by orders of magnitude
compared to the traditional online red-teaming pipeline.
This reduction arises from eliminating iterative generation and multi-model
interaction, thereby removing the primary scalability bottleneck of conventional
red-teaming approaches.

\begin{algorithm*}[htbp]
\caption{JSS Core Computation for Model $M$, For clarity, the pseudocode illustrates a single run on the full dataset. In our experiments, we wrap this procedure in a bootstrap loop to obtain empirical confidence intervals and perform non-parametric statistical tests across models.}
\label{alg:jss_core_calc_en}
\begin{algorithmic}[1]
\REQUIRE 
    Model $M$; datasets $\mathcal{D}_A$ for $A \in \{B,J,R,P\}$; 
    hyperparameters $(r, G, I, s_0, \{w_i\}, \{w_i'\})$.
\ENSURE 
    $\mathrm{JSS}(J,B)$, 
    $\mathrm{JSS}(J,R)$, 
    $\mathrm{JSS}(P,B)$
    .

\STATE \textbf{Trajectory extraction and benign manifold fitting}
\FOR{each $A \in \{B,J,R,P\}$}
    \STATE $\{\Gamma_M^{(A,G)}\}_G \leftarrow \textsc{ExtractAndNormalize}(\mathcal{D}_A, M, G)$
\ENDFOR
\STATE $\mathcal{H}_B^{(G)} \leftarrow$ collect all $h^{(B,G)}$ from $\Gamma_M^{(B,G)}$ for each $G$
\STATE $(\mu^{(G)}, U_r^{(G)}, \Lambda_r^{(G)}) \leftarrow \textsc{WhitenedPCA}(\mathcal{H}_B^{(G)}, r)$

\STATE \textbf{Turning-angle computation}
\FOR{each $A \in \{B,J,R,P\}$ and each $G$}
    \STATE $\Theta^{(A,G)} \leftarrow \textsc{CalculateAngles}(\Gamma_M^{(A,G)}, \mu^{(G)}, U_r^{(G)})$ \COMMENT{all $(\theta_t^{(A,G)}, s_t)$}
\ENDFOR

\STATE \textbf{Slice-wise JS aggregation}
\FOR{each comparison $(A,B') \in \{(J,B),(J,R),(B,R),(P,B)\}$}
    \FOR{each $G$}
        \STATE $\mathrm{JSS}^{(G)}(A,B') \leftarrow 0$, $\mathrm{JSS}^{(G)}_{\text{early}}(A,B') \leftarrow 0$
        \FOR{$i = 1$ to $I$}
            \STATE $\Theta_A^{(G)}(s_i) \leftarrow \{\theta: (\theta,s)\in \Theta^{(A,G)}, s\in s_i\}$
            \STATE $\Theta_{B'}^{(G)}(s_i) \leftarrow \{\theta: (\theta,s)\in \Theta^{(B',G)}, s\in s_i\}$
            \STATE $\mathrm{JS}_i^{(G)}(A,B') \leftarrow \mathrm{JS}\bigl(\mathrm{Dist}(\Theta_A^{(G)}(s_i)) \,\big\|\, \mathrm{Dist}(\Theta_{B'}^{(G)}(s_i))\bigr)$
            \STATE $\mathrm{JSS}^{(G)}(A,B') \mathrel{+}= \mathrm{JS}_i^{(G)}(A,B')$
        \ENDFOR
    \ENDFOR
    \STATE $\mathrm{JSS}(A,B') \leftarrow \frac{1}{|G|}\sum_G \mathrm{JSS}^{(G)}(A,B')$
\ENDFOR

\RETURN $\mathrm{JSS}(J,B)$, 
$\mathrm{JSS}(J,R)$, 
$\mathrm{JSS}(P,B)$
\end{algorithmic}
\end{algorithm*}

\begin{table}[t]
\centering
\caption{Temporal lag between JSS changes and Forward DPO surface-level metrics across dialogue families (UR: Unsafe Rate; RR: Refusal Rate). 
\textbf{Lag Prev.} denotes the percentage of trajectories in which JSS leads the corresponding surface metric, and 
\textbf{Median Shift} reports the median temporal offset (in training steps), with negative values indicating that changes in JSS precede changes in the surface metric.}
\label{tab:jss_lag_analysis}
\scalebox{0.75}{%
\begin{tabular}{l l c c}
\hline
\textbf{Metric} & \textbf{Family} & \textbf{Lag Prev.} & \textbf{Median Shift} \\
\hline
JSS vs.\ RR & RefEx--JB      & 74 & $-4.4$ \\
JSS vs.\ RR & JB--Baseline  & 64 & $-1.4$ \\
JSS vs.\ UR & JB--Baseline  & 53 & $-0.6$ \\
\hline
\end{tabular}%
}
\end{table}

\begin{figure*}[t]
  \centering
    \centering
    \begin{subfigure}{0.48\textwidth}
      \includegraphics[width=\textwidth]{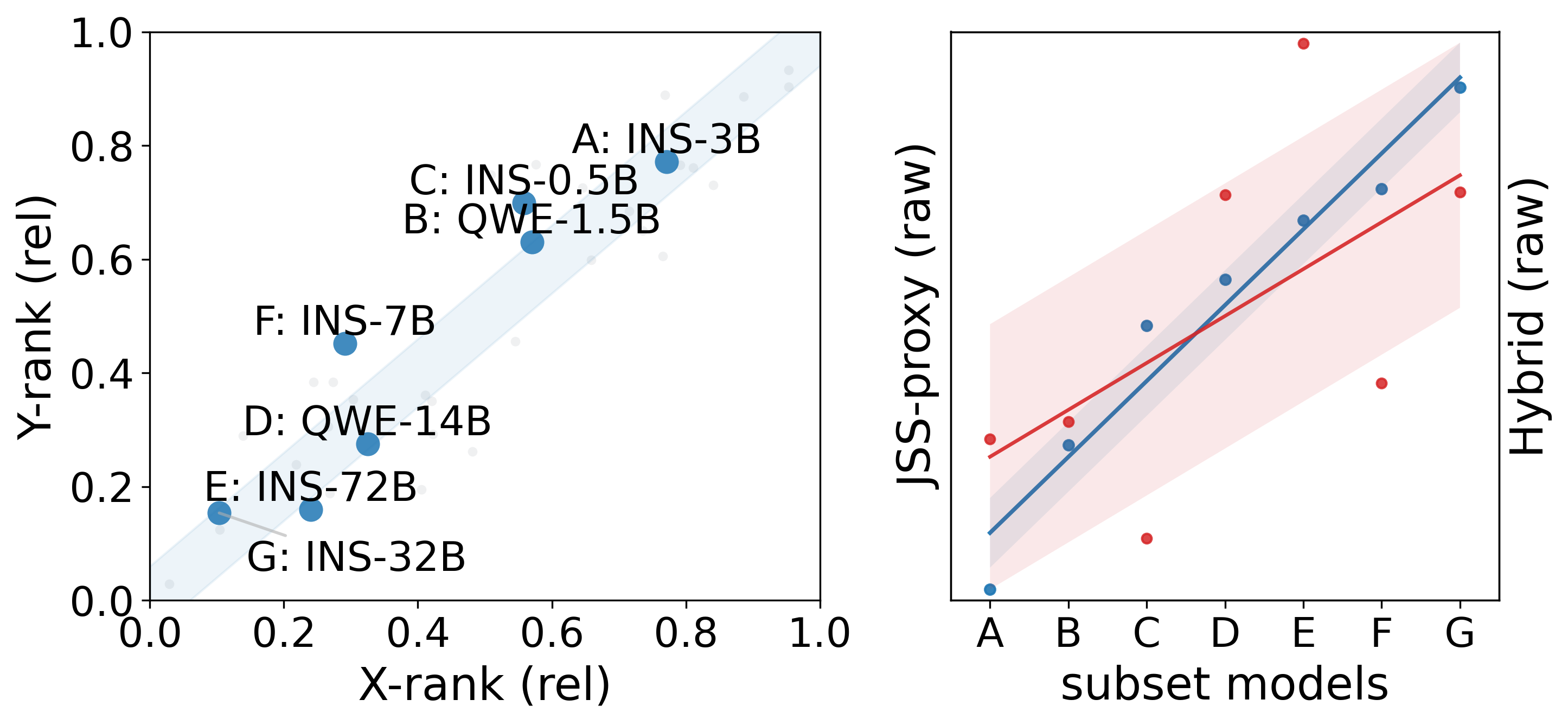}
      \caption{Qwen Family}
    \end{subfigure}
    \hfill
    \begin{subfigure}{0.48\textwidth}
      \includegraphics[width=\textwidth]{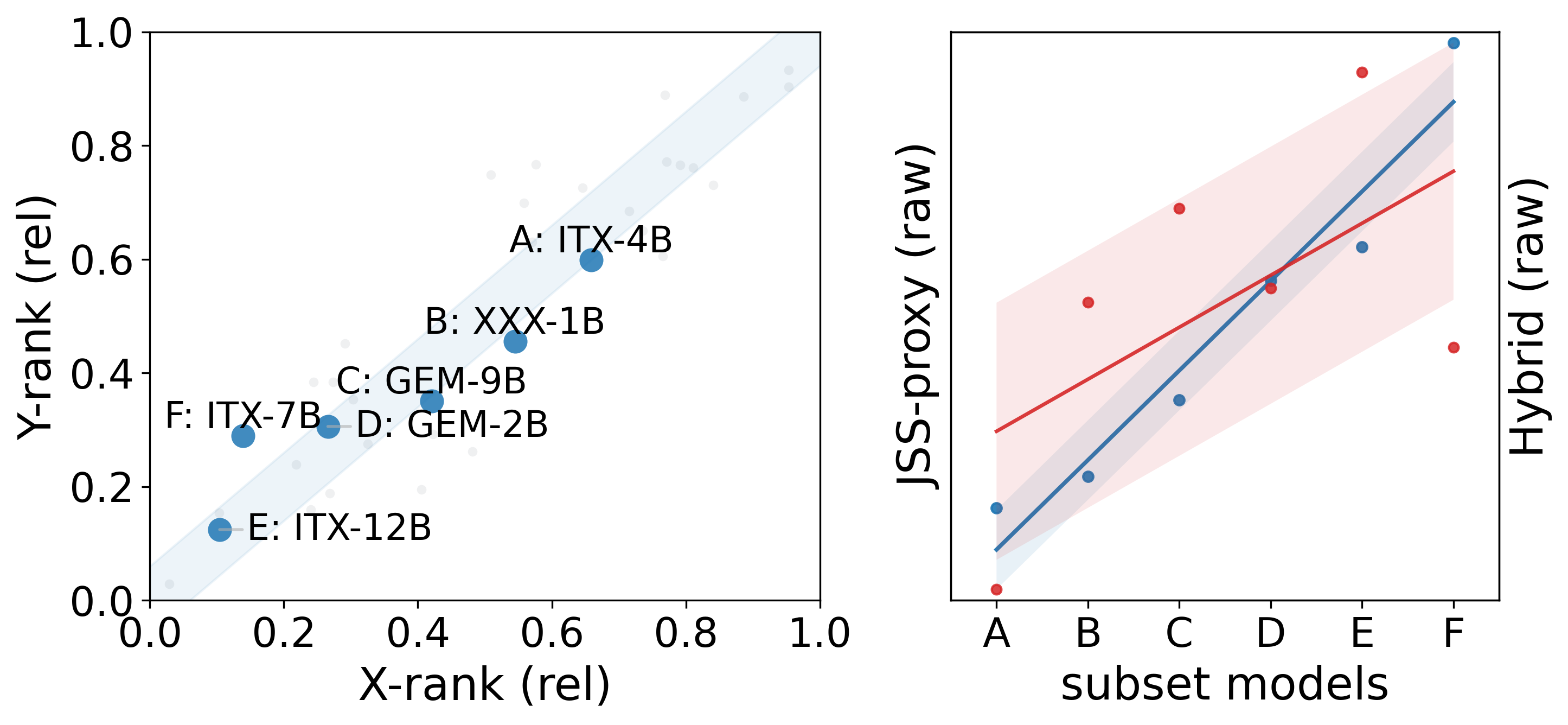}
      \caption{Gemma Family}
    \end{subfigure}

    \begin{subfigure}{0.48\textwidth}
      \includegraphics[width=\textwidth]{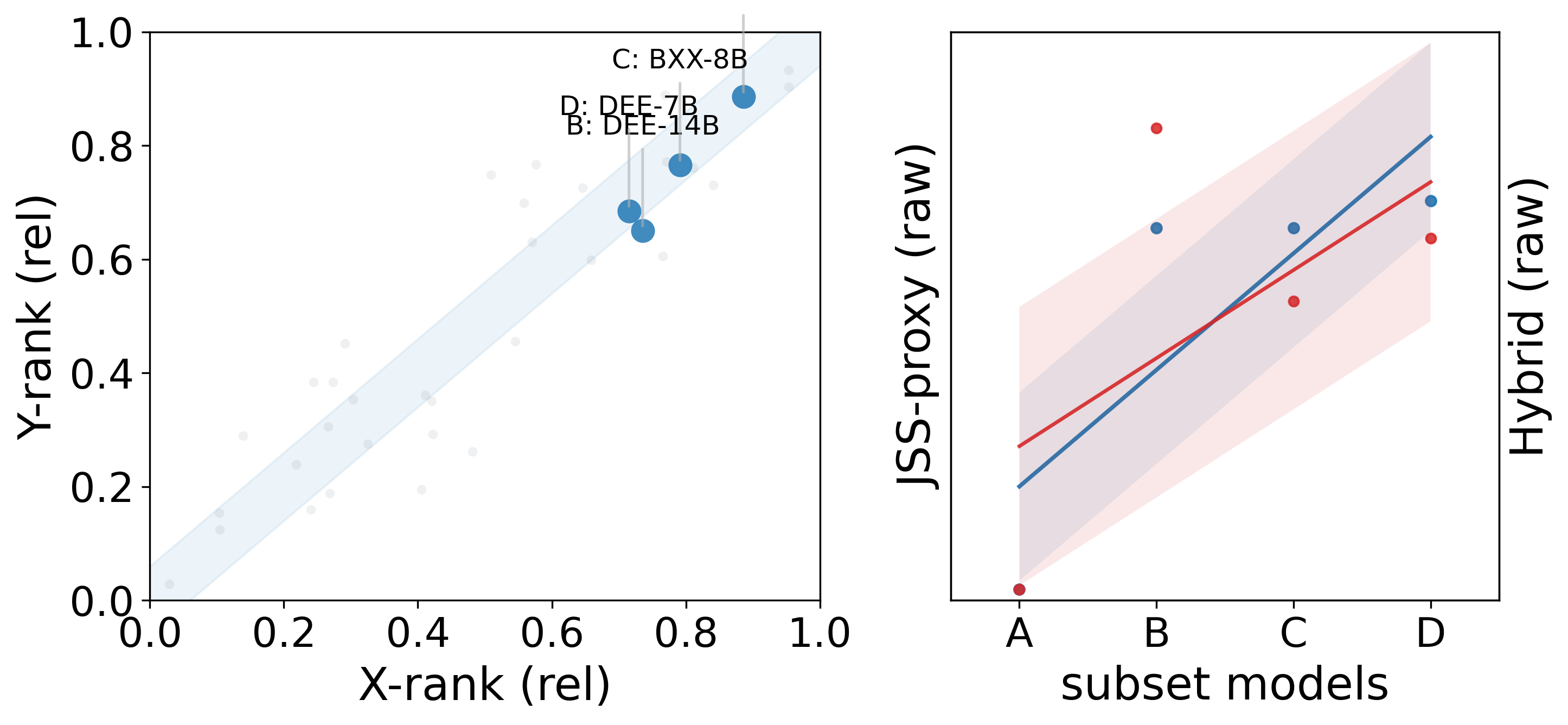}
      \caption{DeepSeek Family}
    \end{subfigure}
    \hfill
    \begin{subfigure}{0.48\textwidth}
      \includegraphics[width=\textwidth]{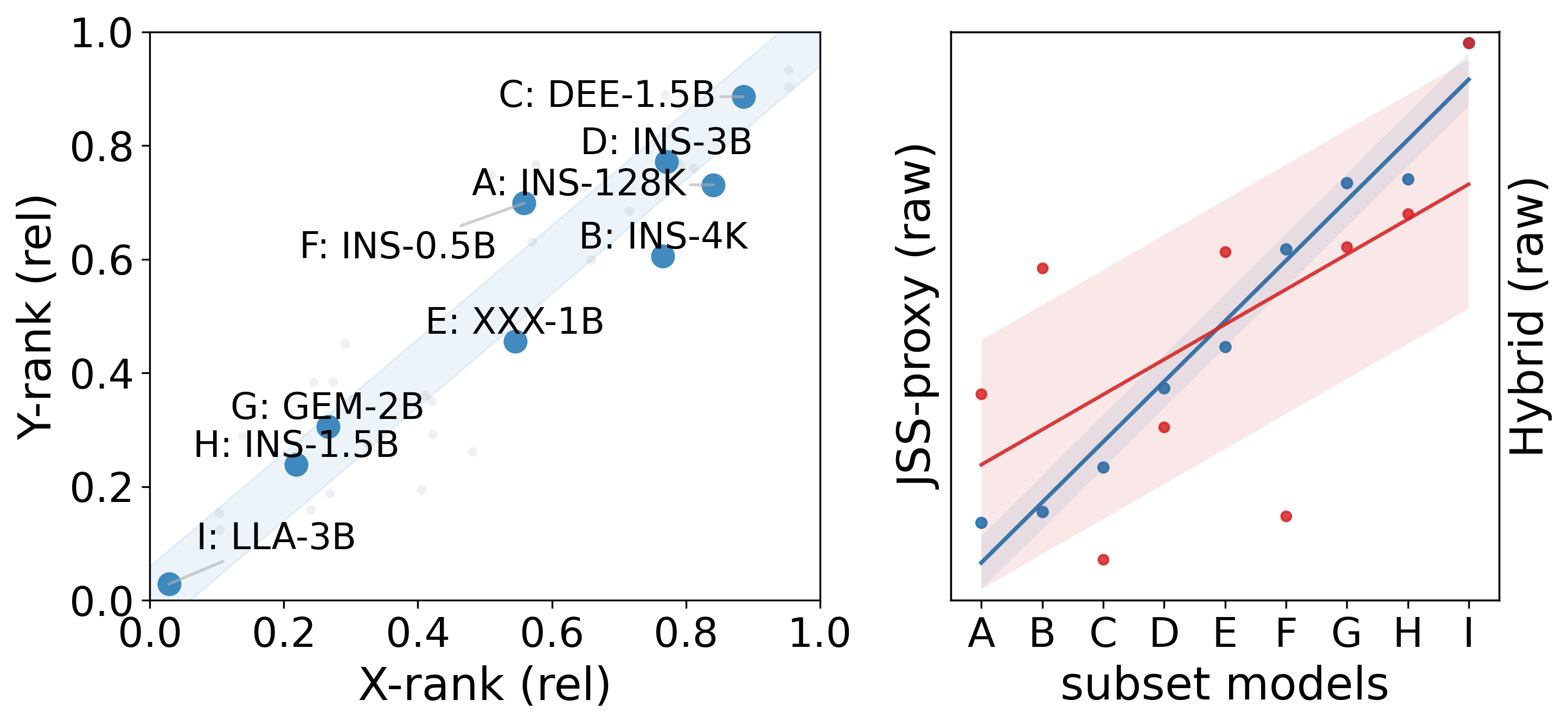}
      \caption{LessEq 3B Family}
    \end{subfigure}
    
    \begin{subfigure}{0.48\textwidth}
      \includegraphics[width=\textwidth]{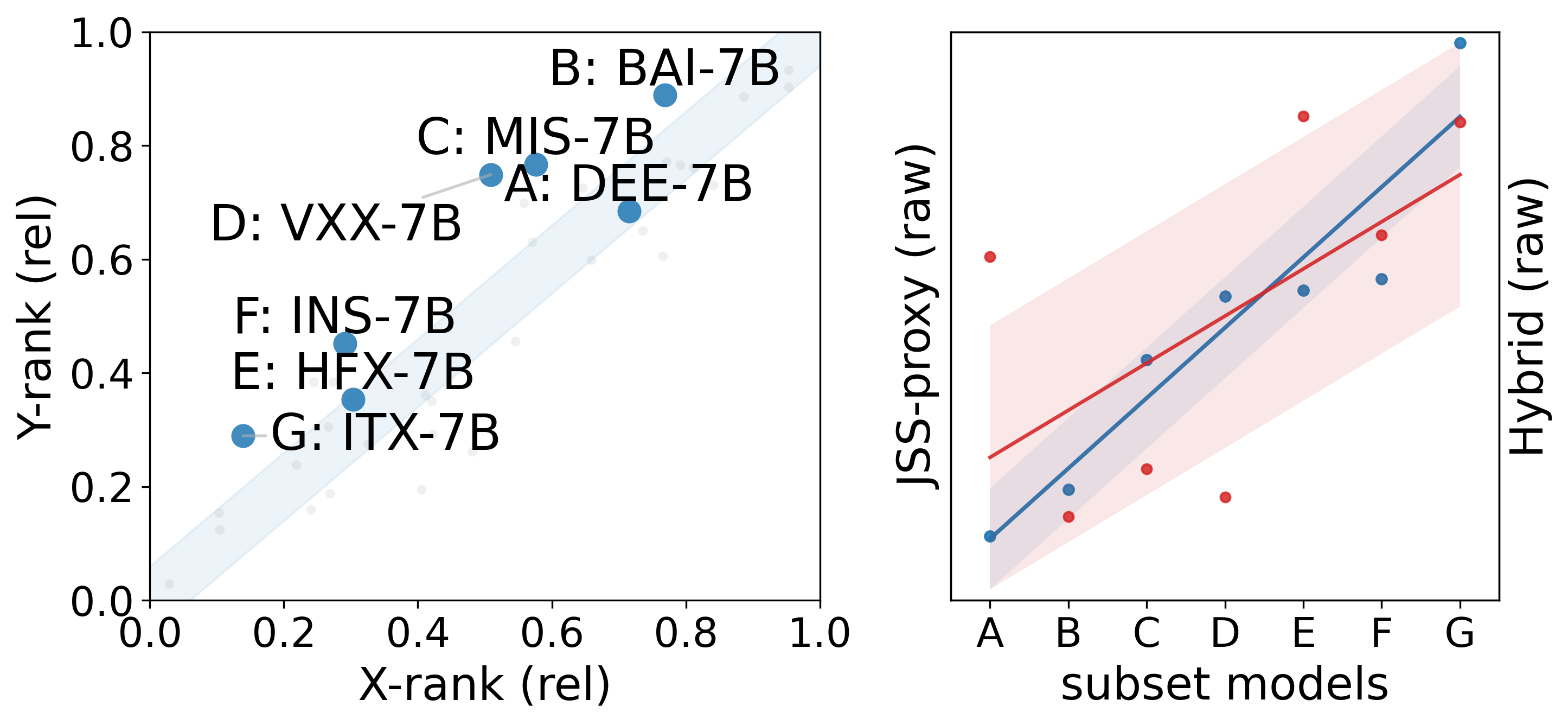}
      \caption{Generic 7B Family}
    \end{subfigure}
    \hfill
    \begin{subfigure}{0.48\textwidth}
      \includegraphics[width=\textwidth]{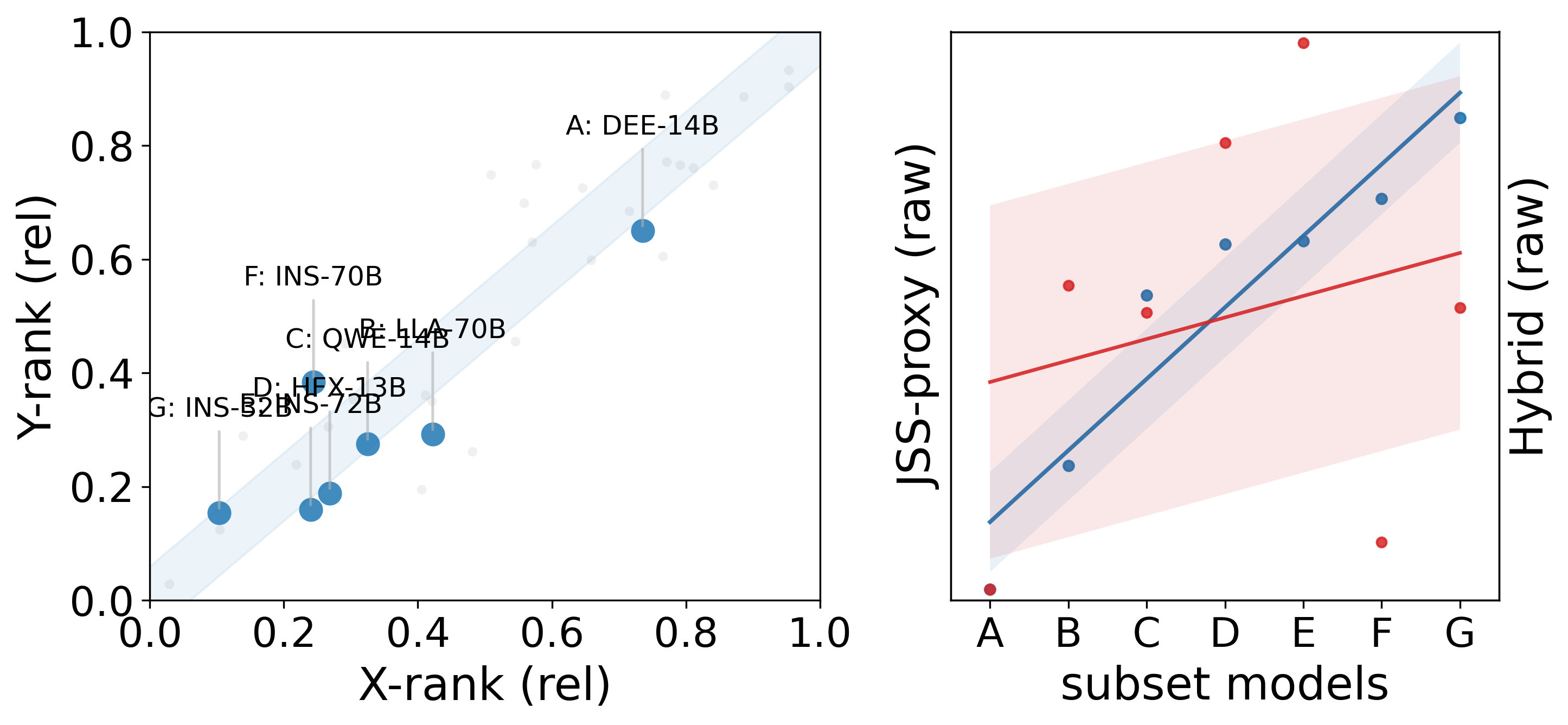}
      \caption{GreatEq 13B Family}
    \end{subfigure}
  
  \caption{\textbf{Multi-perspective cross-model safety ranking consistency between N-GLARE and traditional red-teaming.} For each model family ((a–f; see Appendix Table~\ref{tab:model_abbr_map} for model abbreviations on scatters)), the left scatter plot compares the relative ranks induced by our JSS-based mixed proxy (X-axis) and by Hybrid red-teaming scores (Y-axis); points lying near the diagonal blue band indicate close agreement between the two rankings. The right panel in each family shows the corresponding raw JSS-proxy (blue) and Hybrid (red) scores over all subset models, with shaded regions denoting bootstrap variability, again yielding almost identical within-family orderings. Note that "Hybrid" is the overall average across score all tested red-teaming tasks, with detailed task-wise scores provided in Appendix Table~\ref{tab:jb_safety_benchmark_split_A} and Table~\ref{tab:jb_safety_benchmark_split_B}. }
  \label{fig:combined_ranking_from_diff_LLMfamilies_appendix}
\end{figure*}

\begin{figure*}[t]
  \centering
  \setlength{\tabcolsep}{0pt}
  \renewcommand{\arraystretch}{0}

  \begin{subfigure}[t]{0.49\textwidth}
    \centering
    \includegraphics[width=\linewidth]{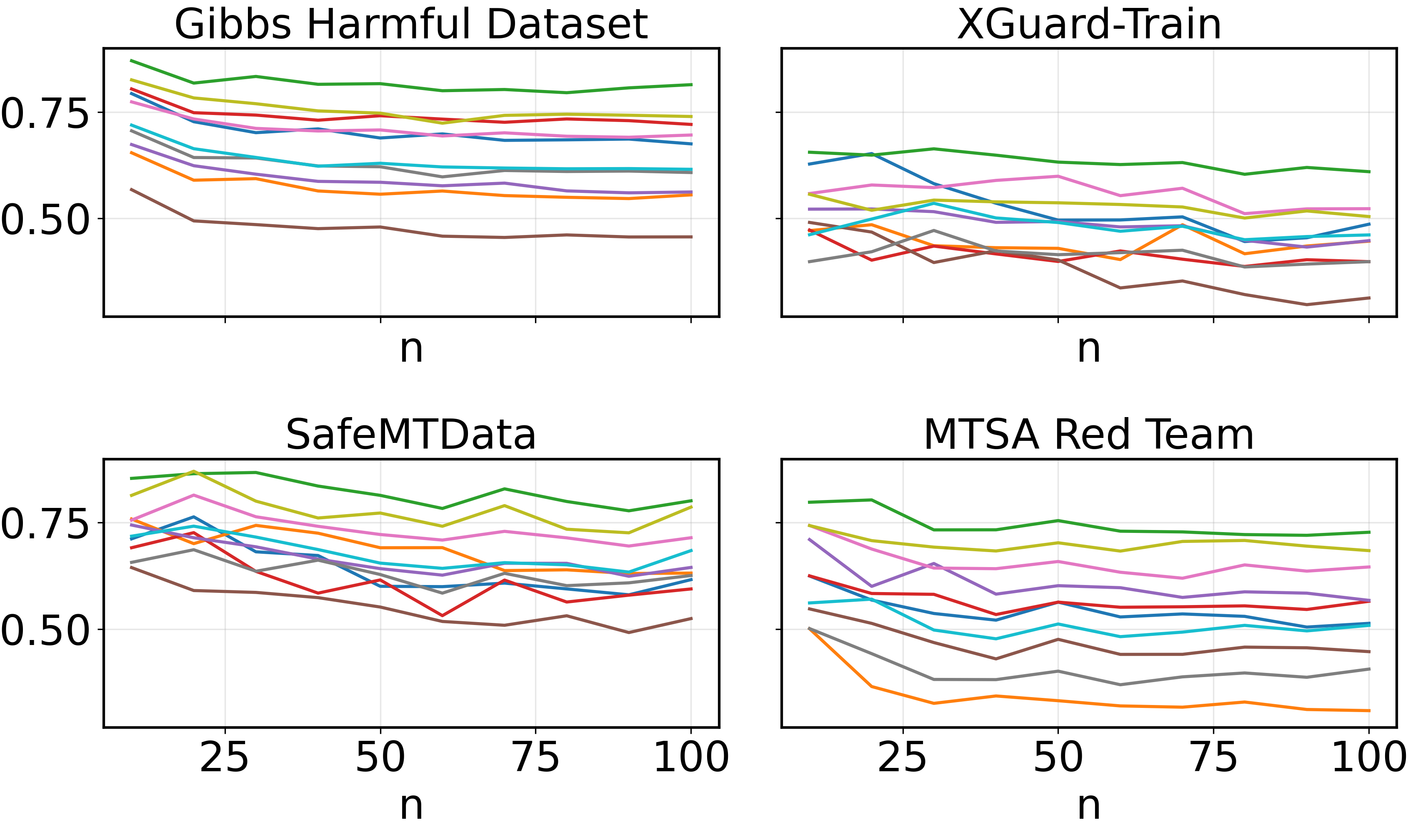}
    \subcaption{Ranking stability across four jailbreak datasets.}\label{fig:big1}
  \end{subfigure}\hfill
  \begin{subfigure}[t]{0.49\textwidth}
    \centering
    \includegraphics[width=\linewidth]{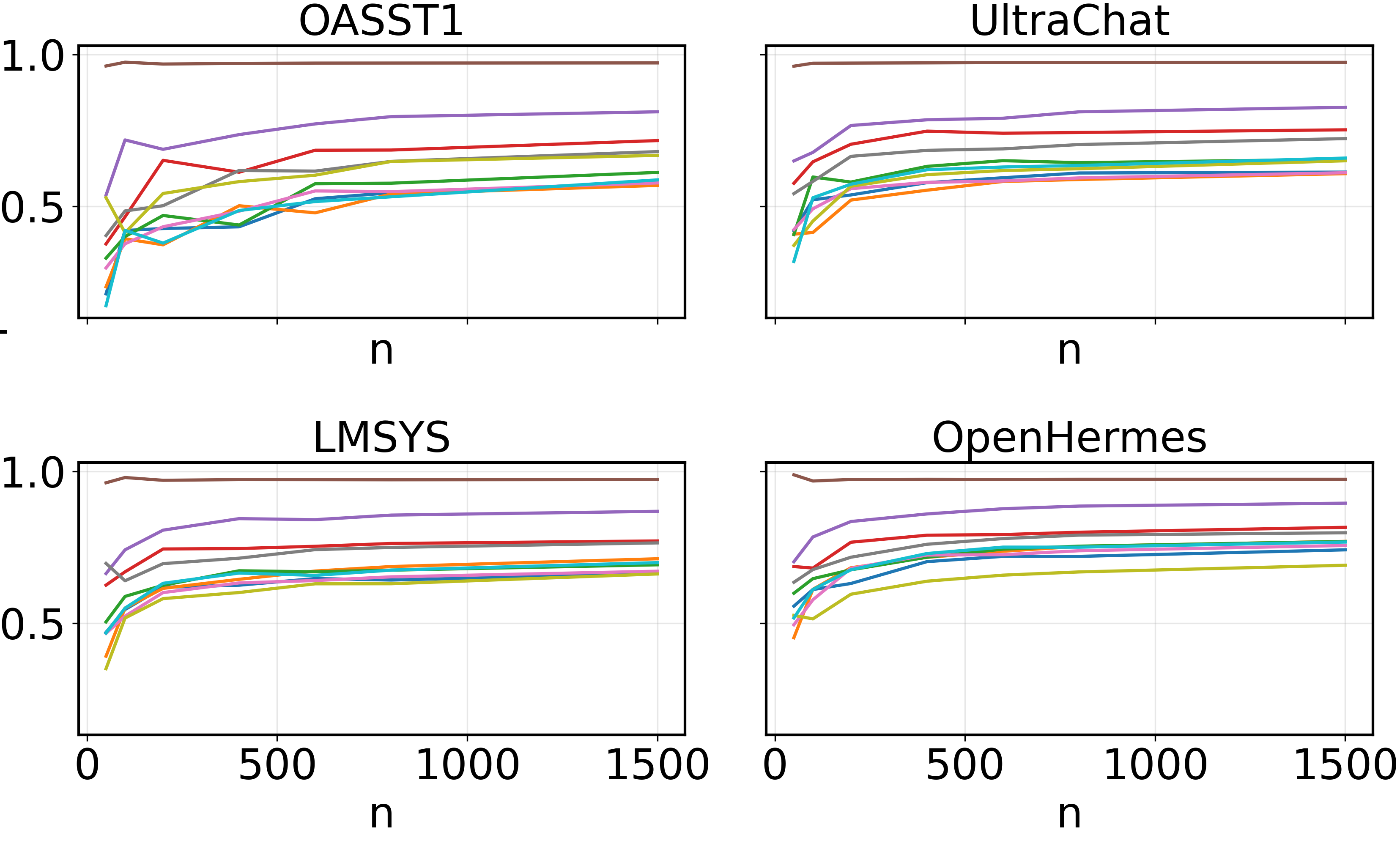}
    \subcaption{Ranking stability across four benign datasets.}\label{fig:big2}
  \end{subfigure}

  \caption{\textbf{Stability of normalized JSS-induced ranking value under varying sample sizes across jailbreak and benign datasets.} Across all settings, the curves remain smooth and largely monotonic, confirming that N-GLARE yields stable safety ordering even when trajectory sampling budgets are perturbed.}
  \label{fig:2x2matrixDataset_only_appendix}
\end{figure*}

\begin{figure*}
    \centering
    \includegraphics[width=1\linewidth]{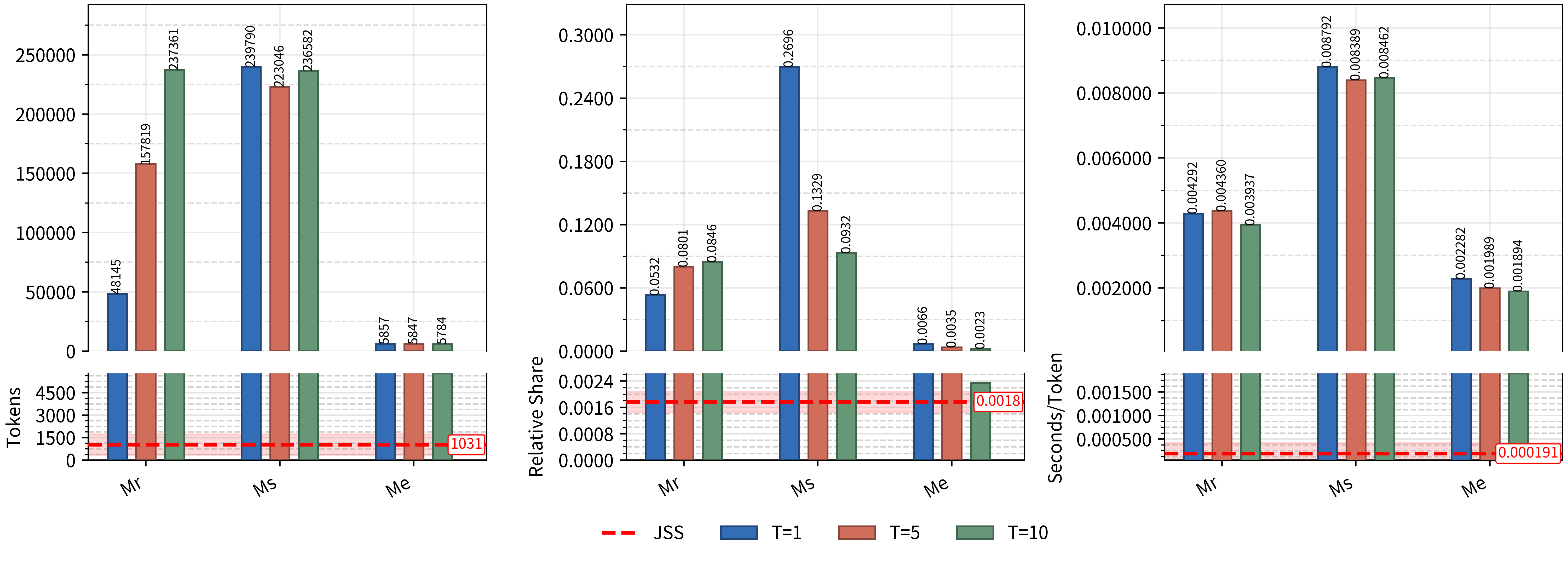}
    \caption{\textbf{Comparative cost breakdown of traditional three-model red-teaming versus our JSS-only pipeline.} For increasing red-teaming depths ($T={1,5,10}$), the three panels report (left) total generated tokens, (middle) the relative share produced by each role—attacker ($M_r$), subject model ($M_s$), evaluator ($M_e$)—and (right) the per-token runtime cost.}
    \label{fig:pyrit_cost_analyze}
\end{figure*}

\begin{figure*}[hbtp]
  \centering
    \scalebox{0.98}{
    \begin{subfigure}[t]{0.48\textwidth}
      \centering
      \includegraphics[width=\linewidth]{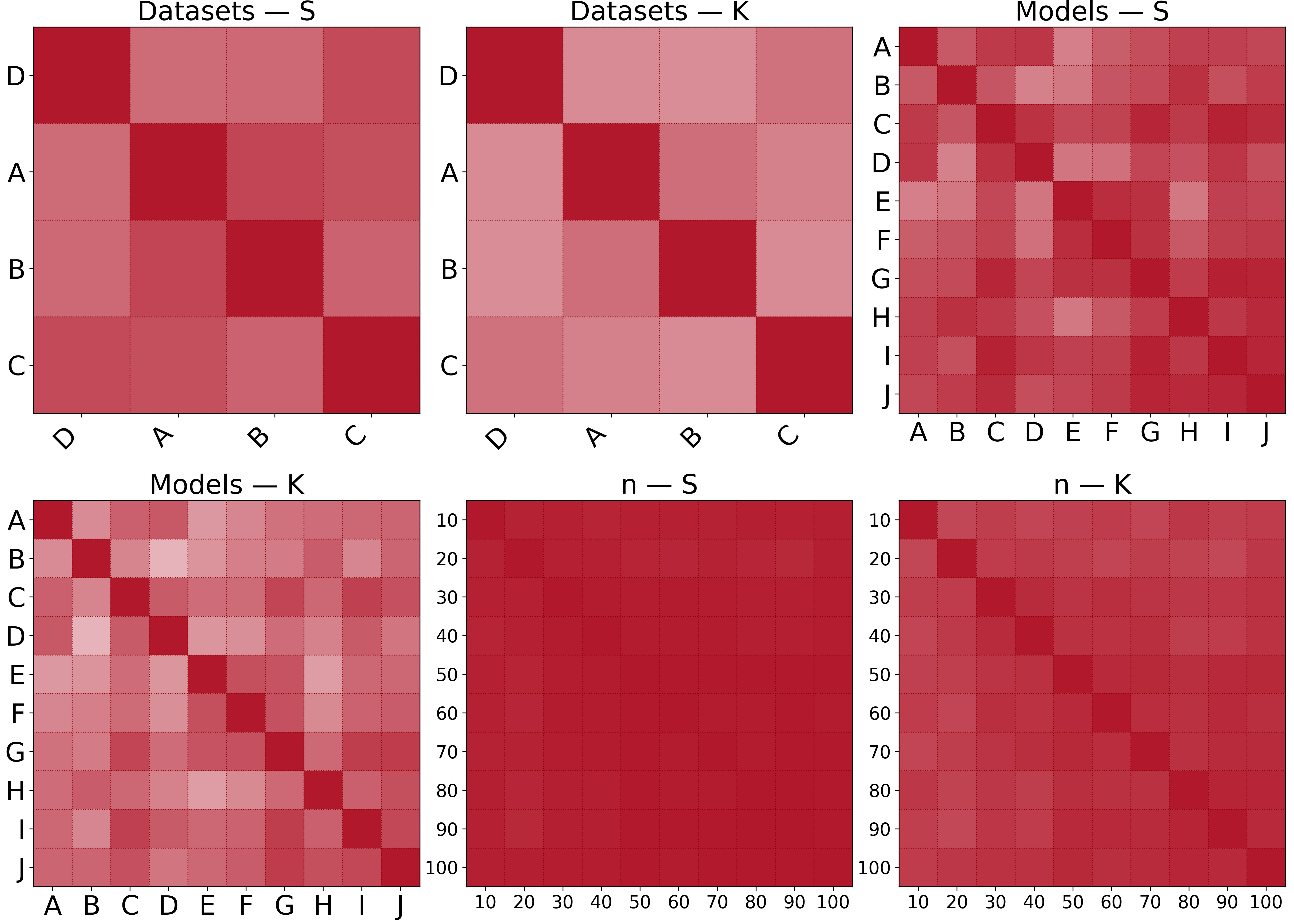}
      \subcaption{Sensitivity analysis on jailbreak datasets.}
      \label{fig:jb_composite_heatmap}
    \end{subfigure}
    }
    \hfill
    \begin{subfigure}[t]{0.48\textwidth}
      \centering
      \includegraphics[width=\linewidth]{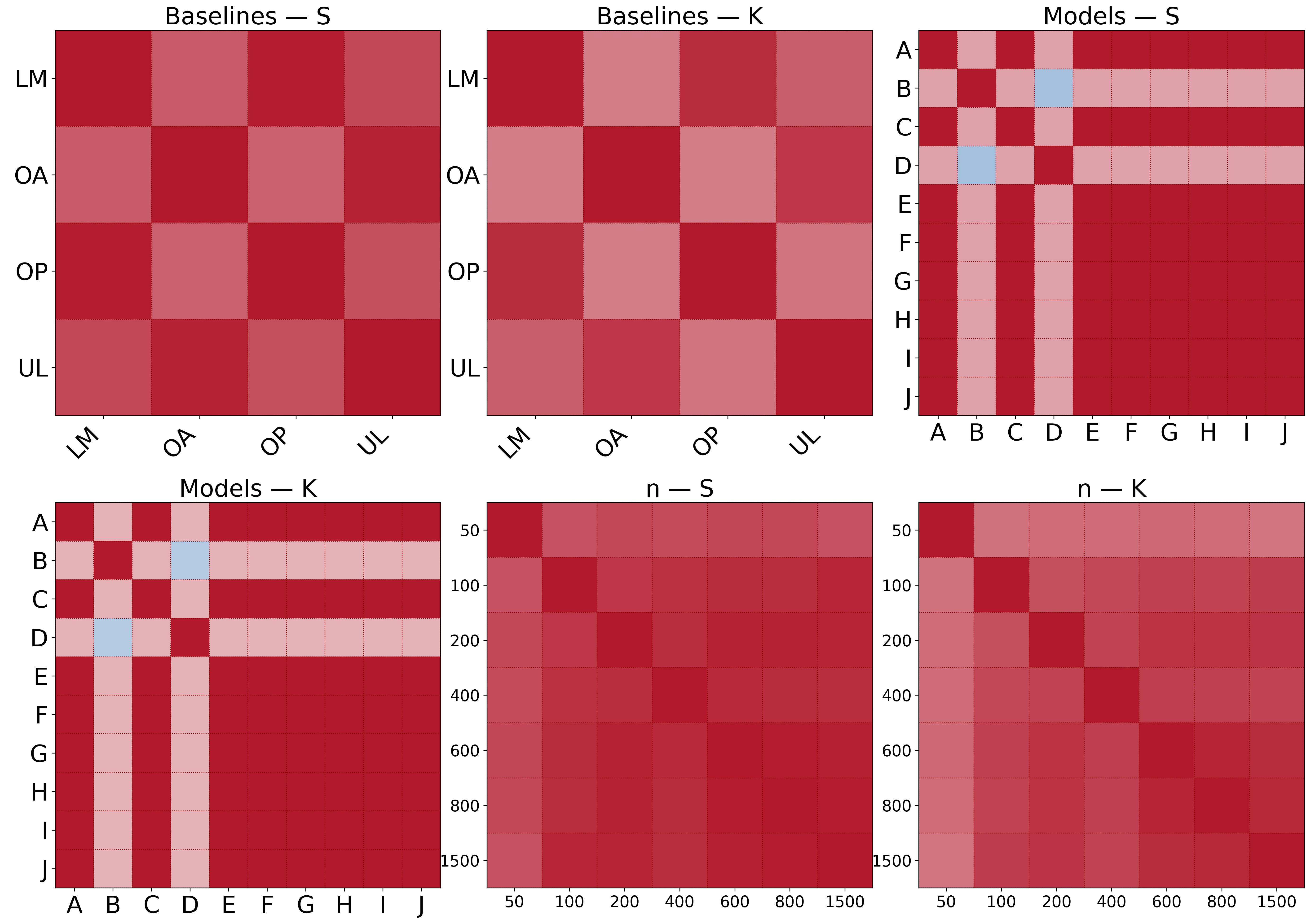}
      \subcaption{Sensitivity analysis on benign datasets.}
      \label{fig:benign_sensitivity_heatmap}
    \end{subfigure}
    \caption{\textbf{Sensitivity heatmaps for JSS-based ranking across datasets, models (see Appendix Table~\ref{tab:modelid_alpha_original} for alphabetical labels mapping), baselines, and sample sizes.} Each heatmap visualizes pairwise rank consistency between two evaluation settings computed with two rank correlation metrics (S = Spearman, K = Kendall), where darker cells indicate higher agreement. Rows and columns correspond to jailbreak datasets, models, baseline datasets, or sampling budgets $n$. For jailbreak analysis, we use four multi-turn attack datasets—(D)=tom-gibbs, (A)=XGuard-Train, (B)=SafeMTData, (C)=MTSARed—and vary the number of included targets to produce small/medium/large (n)-sample conditions (each yielding ($5\text{–}10\times n$) activation trajectories)}
    \label{fig:sensitivity_heatmap}
\end{figure*}

\begin{table}[htbp]
\centering
\small
\caption{Model--Abbreviation Mapping of Subset models Scatter Figure}
\label{tab:model_abbr_map}
\begin{tabular}{l l}
\hline
\textbf{Model} & \textbf{Abbr} \\
\hline
01-ai/Yi-1.5-9B-Chat & CHA-9B \\
Baichuan2-7B-Chat & BAI-7B \\
DeepSeek-R1-Distill-Qwen-1.5B & DEE-1.5B \\
DeepSeek-R1-Distill-Qwen-14B & DEE-14B \\
DeepSeek-R1-Distill-Qwen-7B & DEE-7B \\
Llama-3.1-8B-it & LLA-8B \\
Llama-3.1-Nemotron-70B-Instruct-HF & LLA-70B \\
Llama-3.1-Tulu-3-8B & LLA-8B \\
Llama-3.2-3B-Instruct & LLA-3B \\
Mistral-7B-Instruct-v0.2 & MIS-7B \\
Mistral-NeMo-Minitron-8B-Instruct & MIS-8B \\
Qwen/Qwen2.5-0.5B-Instruct & INS-0.5B \\
Qwen/Qwen2.5-32B-Instruct & INS-32B \\
Qwen/Qwen2.5-3B-Instruct & INS-3B \\
Qwen/Qwen2.5-72B-Instruct & INS-72B \\
Qwen/Qwen2.5-7B-Instruct & INS-7B \\
Qwen2.5-1.5B-Instruct & QWE-1.5B \\
Qwen2.5-14B-Instruct & QWE-14B \\
SakanaAI/TinySwallow-1.5B-Instruct & INS-1.5B \\
Yi-6B-Chat & YIX-6B \\
deepseek-ai/DeepSeek-R1-Distill-Llama-8B & BXX-8B \\
gemma-2-2b-it & GEM-2B \\
gemma-2-9b-it & GEM-9B \\
google/gemma-3-12b-it & ITX-12B \\
google/gemma-3-1b-it & ITX-1B \\
google/gemma-3-4b-it & ITX-4B \\
google/gemma-3-7b-it & ITX-7B \\
granite-3.0-8b-instruct & GRA-8B \\
meta-llama/Llama-2-13b-chat-hf & HFX-13B \\
meta-llama/Llama-2-7b-chat-hf & HFX-7B \\
meta-llama/Meta-Llama-3-70B-Instruct & INS-70B \\
meta-llama/Meta-Llama-3-8B-Instruct & INS-8B \\
microsoft/Phi-3-mini-128k-instruct & INS-128K \\
microsoft/Phi-3-mini-4k-instruct & INS-4K \\
mistralai/Mistral-7B-Instruct-v0.3 & VXX-7B \\
nvidia/Nemotron-Mini-4B-Instruct & INS-4B \\
\hline
\end{tabular}
\end{table}

\begin{table}[htbp]
\centering
\small
\caption{Model ID Mapping with Alphabetical Labels}
\label{tab:modelid_alpha_original}
\begin{tabular}{l c}
\hline
\textbf{Model ID} & \textbf{Letter} \\
\hline
Llama3-8B & A \\
gemma3-4b & B \\
gemma2-9b & C \\
qwen3-8b & D \\
qwen2.5-7b & E \\
deepseek-llama8b & F \\
mistral-7bv3 & G \\
synlogic-7b & H \\
deepseek-qwen1.5b & I \\
Llama2-7B & J \\
\hline
\end{tabular}
\end{table}

\setlength{\tabcolsep}{0.2pt}
\label{langtab}
\onecolumn
\begin{scriptsize}

\begingroup
\setlength{\tabcolsep}{1.5pt}
\renewcommand{\arraystretch}{1.05}
\begin{longtable}{c|p{2.6cm}|*{3}{c}|*{11}{c}}
\caption{Model Safety Benchmark Results (split table A: None to DrAttack).}\label{tab:jb_safety_benchmark_split_A}\\
\hline
 & models & JB/RE & BJ/BP & AI Safety & None & ABJ & Adaptive & ArtPrompt & AutoDAN & Cipher & DAN & DeepInc & DevMode & DRA & DrAttack \\
\hline
\endfirsthead

\hline
 & models & JB/RE & BJ/BP & AI Safety & None & ABJ & Adaptive & ArtPrompt & AutoDAN & Cipher & DAN & DeepInc & DevMode & DRA & DrAttack \\
\hline
\endhead

\hline
\multicolumn{16}{r}{Continued on next page}\\
\endfoot

\hline
\endlastfoot

1 & Llama-3.2-3B-Instruct & 0.6733 & 0.2899 & 81.89 & 91.18 & 83.62 & 44.42 & 47.31 & 97.85 & 100 & 99.57 & 94.40 & 96.12 & 94.05 & 91.38 \\
2 & Qwen/Qwen2.5-7B-Instruct & 0.5346 & 0.4414 & 55.17 & 97.78 & 17.42 & 13.82 & 50.28 & 56.94 & 28.93 & 95.26 & 42.81 & 96.88 & 8.29 & 33.69 \\
3 & mistralai/Mistral-7B-Instruct-v0.3 & 0.3582 & 0.1776 & 34.69 & 61.03 & 8.12 & 6.03 & 25.53 & 47.01 & 3.45 & 20.26 & 25.43 & 23.71 & 2.05 & 31.19 \\
4 & Llama-3.1-8B-it & 0.4312 & 0.2538 & 93.22 & 100 & 99.14 & 65.38 & 100 & 98.71 & 99.43 & 99.14 & 97.63 & 100 & 97.43 & 89.66 \\
5 & google/gemma-7b-it & 0.6992 & 0.5768 & 64.02 & 94.50 & 56.03 & 36.85 & 48.08 & 69.83 & 100 & 72.41 & 38.36 & 97.86 & 19.86 & 71.92 \\
6 & Mistral-7B-Instruct-v0.2 & 0.3400 & 0.3526 & 36.92 & 40.20 & 7.66 & 4.77 & 54.74 & 42.49 & 6.03 & 12.07 & 19.83 & 24.57 & 1.49 & 40.73 \\
7 & nvidia/Nemotron-Mini-4B-Instruct & 0.5784 & 0.3118 & 51.57 & 96.85 & 22.83 & 12.54 & 95.28 & 50.00 & 99.21 & 29.13 & 35.43 & 91.49 & 7.31 & 38.12 \\
8 & Baichuan2-7B-Chat & 0.2792 & 0.5432 & 33.17 & 46.17 & 35.04 & 11.50 & 13.79 & 35.97 & 20.16 & 13.96 & 10.84 & 37.72 & 10.03 & 11.74 \\
9 & gemma-2-9b-it & 0.3329 & 0.4727 & 67.63 & 95.62 & 68.10 & 2.15 & 89.22 & 78.02 & 25.00 & 25.43 & 64.67 & 53.45 & 31.21 & 93.10 \\
10 & TinySwallow-1.5B-Instruct & 0.6708 & 0.3339 & 68.22 & 98.43 & 44.88 & 14.76 & 98.43 & 71.65 & 100 & 94.49 & 44.88 & 100 & 46.13 & 63.24 \\
11 & microsoft/Phi-3-mini-4k-instruct & 0.2995 & 0.5432 & 63.90 & 96.05 & 78.74 & 21.09 & 92.91 & 74.80 & 97.64 & 99.21 & 37.80 & 99.21 & 11.49 & 33.89 \\
12 & google/gemma-3-12b-it & 0.6629 & 0.3645 & 71.16 & 98.45 & 85.83 & 18.11 & 85.04 & 90.55 & 21.71 & 88.28 & 88.19 & 86.99 & 10.51 & 64.66 \\
13 & Llama-3.1-Tulu-3-8B & 0.9034 & 0.2734 & 65.95 & 96.26 & 32.81 & 43.19 & 49.68 & 62.89 & 96.25 & 100 & 38.58 & 98.43 & 10.79 & 56.41 \\
14 & granite-3.0-8b-instruct & 0.4594 & 0.5180 & 61.30 & 99.63 & 27.59 & 58.79 & 28.63 & 63.39 & 55.17 & 89.37 & 15.19 & 91.81 & 12.89 & 36.07 \\
15 & DeepSeek-R1-Distill-Qwen-7B & 0.5685 & 1.0285 & 53.49 & 65.73 & 25.20 & 28.38 & 82.67 & 75.78 & 26.77 & 83.43 & 64.84 & 83.12 & 33.90 & 49.57 \\
16 & microsoft/Phi-3-mini-128k-instruct & 0.2784 & 0.4763 & 53.82 & 94.87 & 30.60 & 6.78 & 84.48 & 71.98 & 80.17 & 95.69 & 21.98 & 97.41 & 5.38 & 26.61 \\
17 & Mistral-NeMo-Minitron-8B-Instruct & 0.7369 & 0.3435 & 47.88 & 84.85 & 19.69 & 21.76 & 35.55 & 48.41 & 14.61 & 43.51 & 29.69 & 36.32 & 3.75 & 43.34 \\
18 & 01-ai/Yi-1.5-9B-Chat & 0.1992 & 0.7507 & 39.36 & 85.17 & 12.93 & 3.39 & 39.24 & 61.64 & 27.35 & 9.83 & 21.98 & 53.06 & 11.75 & 23.60 \\
19 & Yi-6B-Chat & 0.5100 & 0.3796 & 43.36 & 46.25 & 9.32 & 8.78 & 36.01 & 42.34 & 60.51 & 73.18 & 13.26 & 80.69 & 4.87 & 69.47 \\
20 & Qwen/Qwen2.5-0.5B-Instruct & 0.5723 & 0.1777 & 44.05 & 75.06 & 11.71 & 4.45 & 59.99 & 48.15 & 93.13 & 47.45 & 23.24 & 92.67 & 12.81 & 32.20 \\
21 & Qwen2.5-1.5B-Instruct & 0.6726 & 0.3052 & 52.40 & 98.90 & 11.82 & 17.69 & 85.34 & 48.29 & 99.14 & 72.75 & 29.23 & 97.43 & 14.29 & 23.51 \\
22 & Qwen2.5-14B-Instruct & 0.4591 & 0.4041 & 68.61 & 99.63 & 28.80 & 4.27 & 91.38 & 83.19 & 78.45 & 98.28 & 58.80 & 98.28 & 36.62 & 58.19 \\
23 & google/gemma-3-4b-it & 0.4624 & 0.2563 & 57.75 & 93.87 & 50.39 & 11.81 & 56.69 & 75.59 & 17.97 & 42.18 & 69.29 & 39.37 & 22.56 & 49.14 \\
24 & google/gemma-3-1b-it & 0.6095 & 0.5359 & 65.19 & 96.15 & 59.06 & 15.50 & 60.77 & 73.23 & 83.46 & 22.05 & 70.08 & 80.60 & 44.69 & 41.15 \\
25 & gemma-2-2b-it & 0.2755 & 0.2885 & 65.56 & 72.79 & 46.12 & 48.08 & 85.38 & 81.03 & 38.57 & 23.85 & 70.93 & 66.38 & 32.13 & 76.55 \\
26 & DeepSeek-R1-Distill-Qwen-1.5B & 0.2589 & 0.7154 & 40.60 & 41.18 & 19.53 & 30.10 & 61.60 & 82.36 & 27.09 & 31.92 & 66.89 & 15.25 & 17.74 & 49.94 \\
27 & DeepSeek-R1-Distill-Qwen-14B & 0.5352 & 1.4011 & 57.53 & 76.38 & 34.65 & 24.22 & 80.60 & 75.59 & 32.81 & 27.34 & 70.08 & 25.20 & 34.78 & 65.52 \\
28 & deepseek-ai/DeepSeek-R1-Distill-Llama-8B & 0.5352 & 1.4011 & 51.17 & 70.25 & 21.26 & 23.26 & 92.23 & 61.71 & 39.84 & 36.60 & 49.22 & 31.25 & 29.42 & 53.45 \\
29 & meta-llama/Meta-Llama-3-8B-Instruct & 0.0798 & 0.2441 & 79.98 & 91.61 & 71.12 & 80.26 & 72.13 & 97.84 & 100 & 71.37 & 93.99 & 100 & 20.84 & 83.33 \\
30 & meta-llama/Llama-2-7b-chat-hf & 0.7188 & 0.4138 & 64.47 & 46.79 & 98.71 & 7.73 & 99.14 & 88.53 & 100 & 89.80 & 47.09 & 98.71 & 12.33 & 73.07 \\
31 & meta-llama/Llama-2-13b-chat-hf & 0.7386 & 1.4244 & 75.40 & 63.94 & 98.56 & 85.46 & 83.73 & 80.52 & 99.71 & 98.28 & 59.48 & 99.43 & 12.36 & 60.51 \\
32 & Qwen/Qwen2.5-72B-Instruct & 0.5530 & 0.6269 & 79.41 & 97.71 & 54.61 & 10.80 & 98.86 & 78.06 & 100 & 98.58 & 88.67 & 95.74 & 83.13 & 61.76 \\
33 & Qwen/Qwen2.5-32B-Instruct & 0.4748 & 0.5229 & 68.81 & 97.10 & 56.56 & 12.57 & 94.17 & 79.12 & 44.30 & 92.87 & 55.67 & 97.69 & 28.12 & 64.47 \\
34 & Qwen/Qwen2.5-3B-Instruct & 0.3775 & 0.3692 & 51.16 & 98.92 & 23.10 & 9.36 & 36.22 & 49.57 & 63.68 & 81.47 & 25.64 & 97.41 & 11.10 & 26.18 \\
35 & meta-llama/Meta-Llama-3-70B-Instruct & 0.9358 & 0.2999 & 59.43 & 74.28 & 93.10 & 40.14 & 14.35 & 92.75 & 100 & 96.84 & 17.28 & 83.49 & 51.15 & 45.19 \\
36 & Llama-3.1-Nemotron-70B-Instruct-HF & 0.2293 & 0.4283 & 69.69 & 92.96 & 32.28 & 84.15 & 79.31 & 75.78 & 84.37 & 66.93 & 54.33 & 78.74 & 14.30 & 60.34 \\
\end{longtable}
\endgroup

\begingroup
\setlength{\tabcolsep}{1.5pt}
\renewcommand{\arraystretch}{1.05}
\begin{longtable}{c|p{2.6cm}|*{3}{c}|*{10}{c}}
\caption{Model Safety Benchmark Results (split table B: GCG to TAP).}\label{tab:jb_safety_benchmark_split_B}\\
\hline
 & models & JB/RE & BJ/BP & AI Safety & GCG & GPTFuzzer & Grandmother & Masterkey & MultiLing & PAIR & PastTense & Psychology & ReNeLLM & TAP \\
\hline
\endfirsthead

\hline
 & models & JB/RE & BJ/BP & AI Safety & GCG & GPTFuzzer & Grandmother & Masterkey & MultiLing & PAIR & PastTense & Psychology & ReNeLLM & TAP \\
\hline
\endhead

\hline
\multicolumn{15}{r}{Continued on next page}\\
\endfoot

\hline
\endlastfoot

1 & Llama-3.2-3B-Instruct & 0.6733 & 0.2899 & 81.89 & 95.72 & 90.95 & 97.71 & 98.28 & 78.83 & 97.41 & 100 & 95.69 & 64.08 & 98.28 \\
2 & Qwen/Qwen2.5-7B-Instruct & 0.5346 & 0.4414 & 55.17 & 86.80 & 33.16 & 90.24 & 98.63 & 95.09 & 86.51 & 99.52 & 93.10 & 33.91 & 84.74 \\
3 & mistralai/Mistral-7B-Instruct-v0.3 & 0.3582 & 0.1776 & 34.69 & 37.87 & 23.62 & 70.69 & 71.55 & 63.12 & 79.05 & 84.05 & 80.60 & 23.41 & 78.02 \\
4 & Llama-3.1-8B-it & 0.4312 & 0.2538 & 93.22 & 99.57 & 95.30 & 98.28 & 100 & 93.63 & 97.41 & 98.71 & 100 & 90.52 & 100 \\
5 & google/gemma-7b-it & 0.6992 & 0.5768 & 64.02 & 90.95 & 64.39 & 96.12 & 90.95 & 69.26 & 80.60 & 97.84 & 96.12 & 29.70 & 90.09 \\
6 & Mistral-7B-Instruct-v0.2 & 0.3400 & 0.3526 & 36.92 & 29.13 & 32.61 & 63.52 & 78.02 & 60.25 & 65.95 & 97.41 & 79.74 & 32.62 & 67.38 \\
7 & nvidia/Nemotron-Mini-4B-Instruct & 0.5784 & 0.3118 & 51.57 & 93.97 & 33.00 & 85.83 & 99.21 & 92.13 & 82.68 & 96.06 & 74.02 & 23.08 & 75.00 \\
8 & Baichuan2-7B-Chat & 0.2792 & 0.5432 & 33.17 & 23.35 & 45.94 & 71.80 & 47.86 & 58.18 & 74.25 & 84.05 & 72.33 & 24.07 & 62.12 \\
9 & gemma-2-9b-it & 0.3329 & 0.4727 & 67.63 & 96.99 & 24.35 & 85.34 & 100 & 100 & 89.66 & 100 & 99.57 & 78.02 & 92.27 \\
10 & TinySwallow-1.5B-Instruct & 0.6708 & 0.3339 & 68.22 & 98.43 & 57.43 & 96.06 & 98.43 & 97.64 & 86.61 & 100 & 98.43 & 57.14 & 92.13 \\
11 & microsoft/Phi-3-mini-4k-instruct & 0.2995 & 0.5432 & 63.90 & 96.85 & 61.72 & 87.40 & 100 & 68.04 & 88.19 & 100 & 97.64 & 57.98 & 88.19 \\
12 & google/gemma-3-12b-it & 0.6629 & 0.3645 & 71.16 & 96.85 & 63.79 & 88.98 & 97.64 & 100 & 95.28 & 100 & 96.85 & 67.50 & 96.85 \\
13 & Llama-3.1-Tulu-3-8B & 0.9034 & 0.2734 & 65.95 & 88.75 & 73.97 & 100 & 100 & 90.55 & 100 & 100 & 100 & 65.54 & 100 \\
14 & granite-3.0-8b-instruct & 0.4594 & 0.5180 & 61.30 & 98.71 & 60.83 & 99.14 & 99.14 & 78.15 & 91.81 & 99.14 & 98.71 & 61.55 & 86.21 \\
15 & DeepSeek-R1-Distill-Qwen-7B & 0.5685 & 1.0285 & 53.49 & 44.09 & 89.06 & 59.68 & 54.33 & 58.24 & 87.58 & 33.49 & 77.17 & 42.23 & 87.64 \\
16 & microsoft/Phi-3-mini-128k-instruct & 0.2784 & 0.4763 & 53.82 & 93.58 & 47.65 & 80.60 & 100 & 67.42 & 84.48 & 100 & 94.86 & 44.63 & 80.76 \\
17 & Mistral-NeMo-Minitron-8B-Instruct & 0.7369 & 0.3435 & 47.88 & 91.34 & 36.12 & 90.55 & 86.87 & 74.72 & 87.49 & 96.06 & 81.38 & 59.08 & 71.79 \\
18 & 01-ai/Yi-1.5-9B-Chat & 0.1992 & 0.7507 & 39.36 & 67.95 & 27.14 & 74.68 & 86.18 & 47.63 & 77.34 & 93.10 & 75.54 & 27.04 & 76.82 \\
19 & Yi-6B-Chat & 0.5100 & 0.3796 & 43.36 & 74.88 & 31.84 & 81.33 & 94.40 & 61.35 & 72.10 & 97.84 & 92.24 & 23.35 & 67.72 \\
20 & Qwen/Qwen2.5-0.5B-Instruct & 0.5723 & 0.1777 & 44.05 & 61.16 & 38.57 & 80.76 & 67.81 & 86.92 & 76.76 & 98.28 & 61.45 & 23.84 & 66.08 \\
21 & Qwen2.5-1.5B-Instruct & 0.6726 & 0.3052 & 52.40 & 87.17 & 43.02 & 95.69 & 95.69 & 91.10 & 81.03 & 99.14 & 85.78 & 21.97 & 76.06 \\
22 & Qwen2.5-14B-Instruct & 0.4591 & 0.4041 & 68.61 & 93.60 & 67.38 & 96.55 & 100 & 100 & 94.83 & 96.12 & 99.14 & 77.59 & 92.67 \\
23 & google/gemma-3-4b-it & 0.4624 & 0.2563 & 57.75 & 94.53 & 52.13 & 66.93 & 97.64 & 98.43 & 91.34 & 100 & 98.43 & 51.26 & 95.28 \\
24 & google/gemma-3-1b-it & 0.6095 & 0.5359 & 65.19 & 96.85 & 37.07 & 71.65 & 100 & 96.85 & 93.70 & 99.21 & 94.49 & 65.55 & 88.98 \\
25 & gemma-2-2b-it & 0.2755 & 0.2885 & 65.56 & 98.71 & 19.31 & 82.33 & 100 & 99.57 & 90.09 & 66.67 & 99.57 & 83.33 & 90.09 \\
26 & DeepSeek-R1-Distill-Qwen-1.5B & 0.2589 & 0.7154 & 40.60 & 25.39 & 38.26 & 39.14 & 25.30 & 44.60 & 63.69 & 34.37 & 49.74 & 51.52 & 60.61 \\
27 & DeepSeek-R1-Distill-Qwen-14B & 0.5352 & 1.4011 & 57.53 & 59.06 & 74.05 & 74.21 & 44.09 & 80.31 & 84.72 & 77.07 & 88.98 & 63.03 & 83.20 \\
28 & deepseek-ai/DeepSeek-R1-Distill-Llama-8B & 0.5352 & 1.4011 & 51.17 & 51.97 & 86.11 & 69.29 & 59.84 & 62.20 & 86.03 & 55.35 & 85.04 & 39.50 & 79.42 \\
29 & meta-llama/Meta-Llama-3-8B-Instruct & 0.0798 & 0.2441 & 79.98 & 96.57 & 97.13 & 93.16 & 67.24 & 98.71 & 98.71 & 99.57 & 100 & 87.60 & 98.71 \\
30 & meta-llama/Llama-2-7b-chat-hf & 0.7188 & 0.4138 & 64.47 & 97.41 & 58.40 & 91.38 & 100 & 76.60 & 90.09 & 100 & 99.14 & 85.99 & 88.36 \\
31 & meta-llama/Llama-2-13b-chat-hf & 0.7386 & 1.4244 & 75.40 & 96.57 & 66.72 & 99.14 & 99.57 & 75.49 & 91.38 & 99.57 & 100 & 89.66 & 90.99 \\
32 & Qwen/Qwen2.5-72B-Instruct & 0.5530 & 0.6269 & 79.41 & 98.29 & 86.25 & 95.73 & 100 & 98.29 & 87.75 & 99.43 & 98.29 & 81.63 & 92.88 \\
33 & Qwen/Qwen2.5-32B-Instruct & 0.4748 & 0.5229 & 68.81 & 96.24 & 64.71 & 93.44 & 99.60 & 97.81 & 92.24 & 100 & 97.48 & 70.77 & 89.43 \\
34 & Qwen/Qwen2.5-3B-Instruct & 0.3775 & 0.3692 & 51.16 & 94.45 & 28.74 & 93.10 & 99.14 & 91.02 & 78.11 & 100 & 95.26 & 35.22 & 80.60 \\
35 & meta-llama/Meta-Llama-3-70B-Instruct & 0.9358 & 0.2999 & 59.43 & 91.52 & 83.43 & 81.52 & 94.89 & 68.37 & 57.77 & 54.30 & 94.01 & 45.48 & 55.94 \\
36 & Llama-3.1-Nemotron-70B-Instruct-HF & 0.2293 & 0.4283 & 69.69 & 94.83 & 75.86 & 95.28 & 96.06 & 89.06 & 94.53 & 100 & 98.43 & 69.23 & 93.16 \\
\end{longtable}
\endgroup

\begingroup
\setlength{\tabcolsep}{1.5pt}
\renewcommand{\arraystretch}{1.05}
\begin{longtable}{c|p{2.6cm}|cc|*{11}{c}}
\caption{Model Safety Benchmark (split table A). AI Safety is split into Risks and Trials.}\label{tab:jb_safety_split_A}\\
\hline
 & models & Risks & Trials & None & ABJ & Adaptive & ArtPrompt & AutoDAN & Cipher & DAN & DeepInc & DevMode & DRA & DrAttack \\
\hline
\endfirsthead

\hline
 & models & Risks & Trials & None & ABJ & Adaptive & ArtPrompt & AutoDAN & Cipher & DAN & DeepInc & DevMode & DRA & DrAttack \\
\hline
\endhead

\hline
\multicolumn{15}{r}{Continued on next page}\\
\endfoot

\hline
\endlastfoot

1 & Llama-3.2-3B-Instruct & 815 & 5860 & 36/408 & 38/232 & 211/464 & 244/464 & 4/232 & 0/232 & 1/232 & 13/232 & 9/232 & 14/348 & 20/232 \\
2 & Qwen/Qwen2.5-7B-Instruct & 18586 & 68740 & 439/22435 & 2164/2624 & 2260/2624 & 1065/2146 & 1123/2617 & 1776/2517 & 110/2359 & 953/1667 & 62/2145 & 3434/3793 & 1398/2117 \\
3 & mistralai/Mistral-7B-Instruct-v0.3 & 2863 & 4912 & 106/272 & 213/232 & 218/232 & 172/232 & 122/232 & 224/232 & 185/232 & 173/232 & 177/232 & 226/232 & 159/232 \\
4 & Llama-3.1-8B-it & 230 & 5167 & 0/34 & 2/232 & 128/435 & 0/232 & 3/232 & 1/174 & 2/232 & 11/464 & 0/232 & 3/232 & 24/232 \\
5 & google/gemma-7b-it & 1330 & 5028 & 14/272 & 102/232 & 118/232 & 119/232 & 70/232 & 0/232 & 64/232 & 143/232 & 3/232 & 166/232 & 97/348 \\
6 & Mistral-7B-Instruct-v0.2 & 3591 & 6101 & 728/1229 & 214/232 & 220/232 & 105/232 & 133/232 & 218/232 & 204/232 & 186/232 & 175/232 & 228/232 & 275/464 \\
7 & nvidia/Nemotron-Mini-4B-Instruct & 927 & 2601 & 4/127 & 98/127 & 110/127 & 6/127 & 63/127 & 1/127 & 90/127 & 82/127 & 8/127 & 104/116 & 71/116 \\
8 & Baichuan2-7B-Chat & 2853 & 4939 & 145/299 & 150/232 & 198/232 & 199/232 & 122/232 & 182/232 & 198/232 & 205/232 & 138/232 & 178/232 & 183/232 \\
9 & gemma-2-9b-it & 1400 & 4912 & 9/272 & 74/232 & 227/232 & 25/232 & 51/232 & 174/232 & 173/232 & 80/232 & 108/232 & 156/232 & 6/232 \\
10 & TinySwallow-1.5B-Instruct & 525 & 2637 & 2/127 & 70/127 & 92/127 & 2/127 & 36/127 & 0/127 & 7/127 & 70/127 & 0/127 & 65/127 & 42/116 \\
11 & microsoft/Phi-3-mini-4k-instruct & 662 & 2803 & 7/177 & 27/127 & 100/127 & 9/127 & 32/127 & 3/127 & 1/127 & 79/127 & 1/127 & 109/127 & 76/116 \\
12 & google/gemma-3-12b-it & 557 & 2639 & 2/129 & 18/127 & 104/127 & 19/127 & 12/127 & 93/127 & 14/127 & 15/127 & 12/127 & 112/127 & 41/116 \\
13 & Llama-3.1-Tulu-3-8B & 680 & 2887 & 4/159 & 85/127 & 89/159 & 79/159 & 59/159 & 5/159 & 0/127 & 78/127 & 2/127 & 112/127 & 50/116 \\
14 & granite-3.0-8b-instruct & 2595 & 7580 & 1/272 & 252/348 & 325/812 & 165/232 & 166/464 & 156/348 & 37/348 & 295/348 & 19/232 & 390/464 & 218/348 \\
15 & DeepSeek-R1-Distill-Qwen-7B & 979 & 2687 & 60/177 & 95/127 & 85/127 & 14/127 & 30/127 & 93/127 & 11/127 & 44/127 & 18/127 & 56/127 & 58/116 \\
16 & microsoft/Phi-3-mini-128k-instruct & 1581 & 4912 & 13/272 & 161/232 & 216/232 & 36/232 & 65/232 & 46/232 & 10/232 & 181/232 & 6/232 & 218/232 & 170/232 \\
17 & Mistral-NeMo-Minitron-8B-Instruct & 1108 & 2612 & 12/127 & 102/127 & 97/127 & 80/127 & 63/127 & 108/127 & 69/127 & 89/127 & 69/127 & 110/116 & 62/116 \\
18 & 01-ai/Yi-1.5-9B-Chat & 2487 & 4912 & 36/272 & 202/232 & 224/232 & 136/232 & 89/232 & 168/232 & 209/232 & 181/232 & 96/232 & 198/232 & 177/232 \\
19 & Yi-6B-Chat & 2784 & 60676 & 761/1027 & 210/232 & 208/232 & 141/232 & 1126/232 & 91/232 & 60/232 & 200/232 & 44/232 & 212/232 & 68/232 \\
20 & Qwen/Qwen2.5-0.5B-Instruct & 2058 & 4912 & 64/272 & 199/232 & 213/232 & 91/232 & 108/232 & 15/232 & 119/232 & 173/232 & 17/232 & 175/232 & 151/232 \\
21 & Qwen2.5-1.5B-Instruct & 1613 & 4912 & 3/272 & 202/232 & 163/232 & 34/232 & 119/232 & 2/232 & 61/232 & 163/232 & 4/232 & 183/232 & 176/232 \\
22 & Qwen2.5-14B-Instruct & 1025 & 4912 & 1/272 & 164/232 & 222/232 & 20/232 & 39/232 & 50/232 & 3/232 & 95/232 & 3/232 & 144/232 & 97/232 \\
23 & google/gemma-3-4b-it & 995 & 2766 & 6/129 & 63/127 & 112/127 & 55/127 & 31/127 & 104/127 & 73/127 & 39/127 & 77/127 & 191/254 & 59/116 \\
24 & google/gemma-3-1b-it & 738 & 2639 & 4/129 & 52/127 & 107/127 & 46/127 & 34/127 & 21/127 & 99/127 & 38/127 & 23/127 & 59/127 & 67/116 \\
25 & gemma-2-2b-it & 1990 & 6168 & 37/136 & 250/464 & 119/232 & 67/464 & 44/232 & 209/348 & 265/348 & 66/232 & 78/232 & 288/464 & 136/580 \\
26 & DeepSeek-R1-Distill-Qwen-1.5B & 1373 & 2662 & 73/127 & 102/127 & 87/152 & 35/127 & 19/127 & 89/127 & 83/127 & 40/127 & 107/127 & 74/127 & 56/116 \\
27 & DeepSeek-R1-Distill-Qwen-14B & 978 & 2637 & 30/127 & 83/127 & 96/127 & 23/127 & 31/127 & 85/127 & 92/127 & 38/127 & 95/127 & 56/127 & 40/116 \\
28 & deepseek-ai/DeepSeek-R1-Distill-Llama-8B & 1108 & 2783 & 46/157 & 100/127 & 96/127 & 8/127 & 48/127 & 76/127 & 79/127 & 64/127 & 87/127 & 62/127 & 54/116 \\
29 & meta-llama/Meta-Llama-3-8B-Instruct & 844 & 7873 & 167/2798 & 134/464 & 4/29 & 97/348 & 5/232 & 0/232 & 45/290 & 13/232 & 0/232 & 161/232 & 37/232 \\
30 & meta-llama/Llama-2-7b-chat-hf & 3828 & 10893 & 2629/5180 & 3/232 & 24/29 & 1/232 & 23/232 & 0/232 & 20/232 & 293/580 & 3/232 & 271/348 & 93/348 \\
31 & meta-llama/Llama-2-13b-chat-hf & 1749 & 8187 & 868/2503 & 5/348 & 21/232 & 36/348 & 67/348 & 1/348 & 3/232 & 185/464 & 2/348 & 194/232 & 91/232 \\
32 & Qwen/Qwen2.5-72B-Instruct & 1009 & 7497 & 6/391 & 156/351 & 313/351 & 4/351 & 77/351 & 0/351 & 4/351 & 38/351 & 14/351 & 55/351 & 130/340 \\
33 & Qwen/Qwen2.5-32B-Instruct & 7062 & 34002 & 110/4738 & 655/1508 & 1318/1508 & 87/1508 & 314/1504 & 840/1508 & 101/1508 & 668/1508 & 26/1508 & 1143/1624 & 494/1392 \\
34 & Qwen/Qwen2.5-3B-Instruct & 1762 & 4918 & 3/278 & 176/232 & 210/232 & 135/232 & 117/232 & 68/232 & 43/232 & 172/232 & 6/232 & 199/232 & 171/232 \\
35 & meta-llama/Meta-Llama-3-70B-Instruct & 1666 & 6650 & 309/1430 & 16/232 & 139/232 & 199/232 & 17/232 & 0/232 & 7/232 & 192/232 & 38/232 & 113/232 & 127/232 \\
36 & Llama-3.1-Nemotron-70B-Instruct-HF & 537 & 2601 & 8/127 & 86/127 & 11/127 & 26/127 & 31/127 & 20/127 & 42/127 & 58/127 & 27/127 & 109/127 & 50/127 \\
\end{longtable}
\endgroup

\begingroup
\setlength{\tabcolsep}{1.5pt}
\renewcommand{\arraystretch}{1.05}
\begin{longtable}{c|p{2.6cm}|cc|*{10}{c}}
\caption{Model Safety Benchmark (split table B). AI Safety is split into Risks and Trials.}\label{tab:jb_safety_split_B}\\
\hline
 & models & Risks & Trials & GCG & GPTFuzzer & Grandmother & Masterkey & MultiLing & PAIR & PastTense & Psychology & ReNeLLM & TAP \\
\hline
\endfirsthead

\hline
 & models & Risks & Trials & GCG & GPTFuzzer & Grandmother & Masterkey & MultiLing & PAIR & PastTense & Psychology & ReNeLLM & TAP \\
\hline
\endhead

\hline
\multicolumn{14}{r}{Continued on next page}\\
\endfoot

\hline
\endlastfoot

1 & Llama-3.2-3B-Instruct & 815 & 5860 & 8/232 & 21/232 & 7/348 & 4/232 & 40/232 & 6/232 & 0/232 & 10/232 & 125/348 & 4/232 \\
2 & Qwen/Qwen2.5-7B-Instruct & 18586 & 68740 & 2233/1986 & 1014/1537 & 224/2305 & 33/2624 & 88/1986 & 311/2305 & 11/2305 & 159/2305 & 1401/2138 & 338/2215 \\
3 & mistralai/Mistral-7B-Instruct-v0.3 & 2863 & 4912 & 141/232 & 176/232 & 68/232 & 66/232 & 81/232 & 47/232 & 37/232 & 45/232 & 176/232 & 51/232 \\
4 & Llama-3.1-8B-it & 230 & 5167 & 1/232 & 8/232 & 6/348 & 0/232 & 10/232 & 6/232 & 3/232 & 0/232 & 22/232 & 0/232 \\
5 & google/gemma-7b-it & 1330 & 5028 & 21/232 & 80/232 & 9/232 & 21/232 & 61/232 & 45/232 & 5/232 & 9/232 & 160/232 & 23/232 \\
6 & Mistral-7B-Instruct-v0.2 & 3591 & 6101 & 162/232 & 154/232 & 84/232 & 51/232 & 91/232 & 79/232 & 6/232 & 47/232 & 156/232 & 75/232 \\
7 & nvidia/Nemotron-Mini-4B-Instruct & 927 & 2601 & 7/116 & 76/116 & 18/127 & 1/127 & 10/127 & 22/127 & 5/127 & 33/127 & 89/116 & 29/116 \\
8 & Baichuan2-7B-Chat & 2853 & 4939 & 171/232 & 117/232 & 61/232 & 120/232 & 89/232 & 59/232 & 37/232 & 62/232 & 153/232 & 86/232 \\
9 & gemma-2-9b-it & 1400 & 4912 & 6/232 & 174/232 & 34/232 & 0/232 & 0/232 & 24/232 & 0/232 & 1/232 & 51/232 & 17/232 \\
10 & TinySwallow-1.5B-Instruct & 525 & 2637 & 2/127 & 47/116 & 5/127 & 2/127 & 3/127 & 17/127 & 0/127 & 2/127 & 51/119 & 10/127 \\
11 & microsoft/Phi-3-mini-4k-instruct & 662 & 2803 & 4/127 & 81/232 & 16/127 & 0/127 & 34/127 & 15/127 & 0/127 & 3/127 & 50/119 & 15/127 \\
12 & google/gemma-3-12b-it & 557 & 2639 & 4/127 & 42/116 & 14/127 & 3/127 & 0/127 & 6/127 & 0/127 & 4/127 & 38/119 & 4/127 \\
13 & Llama-3.1-Tulu-3-8B & 680 & 2887 & 17/159 & 37/145 & 0/127 & 0/127 & 12/127 & 0/127 & 0/127 & 0/127 & 51/148 & 0/127 \\
14 & granite-3.0-8b-instruct & 2595 & 7580 & 3/232 & 171/464 & 2/232 & 3/348 & 68/348 & 19/232 & 3/348 & 3/232 & 267/696 & 32/232 \\
15 & DeepSeek-R1-Distill-Qwen-7B & 979 & 2687 & 71/127 & 9/116 & 50/127 & 58/127 & 50/127 & 14/127 & 54/127 & 29/127 & 67/119 & 13/127 \\
16 & microsoft/Phi-3-mini-128k-instruct & 1581 & 4912 & 13/232 & 120/232 & 45/232 & 0/232 & 64/232 & 36/232 & 0/232 & 10/232 & 128/232 & 43/232 \\
17 & Mistral-NeMo-Minitron-8B-Instruct & 1108 & 2612 & 11/127 & 71/116 & 12/127 & 14/127 & 24/127 & 15/127 & 5/127 & 22/127 & 41/116 & 32/116 \\
18 & 01-ai/Yi-1.5-9B-Chat & 2487 & 4912 & 68/232 & 165/232 & 58/232 & 24/232 & 116/232 & 51/232 & 16/232 & 56/232 & 164/232 & 53/232 \\
19 & Yi-6B-Chat & 2784 & 60676 & 56/232 & 150/232 & 40/232 & 13/232 & 74/232 & 64/232 & 5/232 & 18/232 & 173/232 & 70/232 \\
20 & Qwen/Qwen2.5-0.5B-Instruct & 2058 & 4912 & 81/232 & 125/232 & 43/232 & 74/232 & 25/232 & 50/232 & 4/232 & 85/232 & 170/232 & 76/232 \\
21 & Qwen2.5-1.5B-Instruct & 1613 & 4912 & 28/232 & 130/232 & 10/232 & 10/232 & 16/232 & 44/232 & 1/232 & 33/232 & 177/232 & 54/232 \\
22 & Qwen2.5-14B-Instruct & 1025 & 4912 & 12/232 & 75/232 & 8/232 & 0/232 & 0/232 & 12/232 & 9/232 & 2/232 & 52/232 & 17/232 \\
23 & google/gemma-3-4b-it & 995 & 2766 & 6/127 & 55/116 & 42/127 & 3/127 & 2/127 & 11/127 & 0/127 & 2/127 & 58/119 & 6/127 \\
24 & google/gemma-3-1b-it & 738 & 2639 & 4/127 & 73/116 & 36/127 & 0/127 & 4/127 & 8/127 & 1/127 & 7/127 & 41/119 & 14/127 \\
25 & gemma-2-2b-it & 1990 & 6168 & 3/232 & 15/232 & 41/232 & 0/232 & 1/232 & 23/232 & 116/348 & 1/232 & 37/232 & 23/232 \\
26 & DeepSeek-R1-Distill-Qwen-1.5B & 1373 & 2662 & 92/127 & 64/116 & 68/127 & 93/127 & 61/127 & 31/127 & 55/127 & 56/127 & 55/119 & 33/127 \\
27 & DeepSeek-R1-Distill-Qwen-14B & 978 & 2637 & 52/127 & 22/116 & 32/127 & 71/127 & 25/127 & 15/127 & 17/127 & 14/127 & 44/119 & 17/127 \\
28 & deepseek-ai/DeepSeek-R1-Distill-Llama-8B & 1108 & 2783 & 61/127 & 26/232 & 39/127 & 51/127 & 48/127 & 16/127 & 34/127 & 19/127 & 92/1190 & 22/127 \\
29 & meta-llama/Meta-Llama-3-8B-Instruct & 844 & 7873 & 6/232 & 10/348 & 14/232 & 114/348 & 3/232 & 3/232 & 1/232 & 0/232 & 27/232 & 3/232 \\
30 & meta-llama/Llama-2-7b-chat-hf & 3828 & 10893 & 6/232 & 283/696 & 20/232 & 0/232 & 42/232 & 23/232 & 0/232 & 2/232 & 65/464 & 27/232 \\
31 & meta-llama/Llama-2-13b-chat-hf & 1749 & 8187 & 6/232 & 149/464 & 2/232 & 1/232 & 53/232 & 20/232 & 1/232 & 0/232 & 24/232 & 20/232 \\
32 & Qwen/Qwen2.5-72B-Instruct & 1009 & 7497 & 6/351 & 45/340 & 15/351 & 0/351 & 6/351 & 43/351 & 1/351 & 8/467 & 63/343 & 25/351 \\
33 & Qwen/Qwen2.5-32B-Instruct & 7062 & 34002 & 50/1508 & 488/1392 & 99/1508 & 6/1508 & 31/1508 & 117/1508 & 0/1189 & 38/1508 & 324/11111 & 153/1448 \\
34 & Qwen/Qwen2.5-3B-Instruct & 1762 & 4918 & 10/232 & 162/232 & 16/232 & 2/232 & 19/232 & 50/232 & 0/232 & 11/232 & 147/232 & 45/232 \\
35 & meta-llama/Meta-Llama-3-70B-Instruct & 1666 & 6650 & 20/232 & 38/232 & 43/232 & 12/232 & 73/232 & 98/232 & 106/232 & 14/232 & 126/232 & 102/232 \\
36 & Llama-3.1-Nemotron-70B-Instruct-HF & 537 & 2601 & 7/127 & 31/127 & 6/127 & 5/127 & 14/127 & 7/127 & 0/127 & 2/127 & 39/127 & 9/127 \\
\end{longtable}
\endgroup

\end{scriptsize}
\end{document}